\documentclass[graybox]{svmult}

\usepackage{type1cm}        % activate if the above 3 fonts are
\usepackage{makeidx}         % allows index generation
\usepackage{graphicx}        % standard LaTeX graphics tool
\usepackage{multicol}        % used for the two-column index
\usepackage[bottom]{footmisc}% places footnotes at page bottom

\usepackage{algorithm}
\usepackage[noend]{algpseudocode}
\makeatletter
\def\BState{\State\hskip-\ALG@thistlm}
\makeatother

\usepackage{newtxtext}       % 
\usepackage{newtxmath}       % selects Times Roman as basic font

\usepackage{times}
\usepackage{geometry}
\usepackage{graphicx}
\usepackage{color}
\usepackage{xcolor}
\usepackage{amssymb}
\usepackage{booktabs}
\usepackage{tabularx}
\usepackage{multirow}
\usepackage{amsmath}
\usepackage{acronym}
\usepackage{tikz}
\usepackage{pstricks}
\usepackage{ifplatform}
\usepackage{xkeyval}
\usepackage{psfrag}
\usepackage{authblk}
\usepackage[numbers]{natbib}
\usepackage{multicol}
\usepackage{hyperref}

\usepackage[format=plain, labelfont={bf,it}, textfont=it]{caption}
\makeindex             % used for the subject index
\graphicspath{{images/}}

\acrodef{ML}{machine learning}

\usepackage{xspace}

\begin{document}

\title*{The AI Driving Olympics at NeurIPS 2018}
\author{Julian Zilly${}^1$, Jacopo Tani${}^1$,  Breandan Considine${}^2$, Bhairav Mehta${}^2$,   Andrea F.\ Daniele${}^{4}$, Manfred Diaz${}^2$, Gianmarco Bernasconi${}^1$, Claudio	Ruch${}^1$, Jan Hakenberg${}^1$, Florian Golemo${}^{2}$, A. Kirsten Bowser${}^5$, Matthew R.\ Walter${}^4$, Ruslan Hristov${}^3$, Sunil Mallya${}^6$, Emilio Frazzoli${}^{1,3}$, Andrea Censi${}^{1,3}$, Liam Paull${}^{2}$ }
\institute{${}^1$ETH Z\"urich, Z\"urich, Switzerland \\ ${}^{2}$Mila, Universit\'e de Montr\'eal, Montr\'eal, Canada\\  
${}^3$nuTonomy, an Aptiv company, Boston, USA \\
${}^4$Toyota Technological Institute at Chicago, Chicago, IL, USA \\
${}^5$Duckietown Foundation, Boston, MA, USA
\\
${}^6$Amazon Web Services, San Francisco, USA}
\maketitle
\vspace{-3cm}
\begin{center}
    \includegraphics[width=0.2\textwidth]{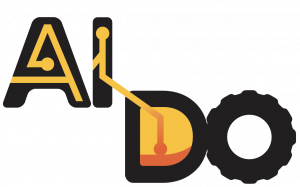}
\end{center}

\abstract*{Despite recent breakthroughs, the ability of deep learning and reinforcement learning to outperform traditional approaches to control physically embodied robotic agents remains largely unproven. 
To help bridge this gap, we created the ``AI Driving Olympics'' (AI-DO), a competition with the objective of evaluating the state of the art in machine learning and artificial intelligence for mobile robotics. 
Based on the simple and well specified autonomous driving and navigation environment called ``Duckietown,'' AI-DO includes a series of tasks of increasing complexity -- from simple lane-following to fleet management. For each task, we provide tools for competitors to use in the form of simulators, logs, code templates, baseline implementations and low-cost access to robotic hardware. We evaluate submissions in simulation online, on standardized hardware environments, and finally at the competition event. The first AI-DO, AI-DO 1, occurred at the Neural Information Processing Systems (NeurIPS) conference in December 2018.
The results of AI-DO 1 highlight the need for better benchmarks, which are lacking in robotics, as well as improved mechanisms to bridge the gap between simulation and reality.
}

\abstract{Despite recent breakthroughs, the ability of deep learning and reinforcement learning to outperform traditional approaches to control physically embodied robotic agents remains largely unproven.  
To help bridge this gap, we created the ``AI Driving Olympics'' (AI-DO), a competition with the objective of evaluating the state of the art in machine learning and artificial intelligence for mobile robotics. 
Based on the simple and well specified autonomous driving and navigation environment called ``Duckietown,'' AI-DO includes a series of tasks of increasing complexity -- from simple lane-following to fleet management. For each task, we provide tools for competitors to use in the form of simulators, logs, code templates, baseline implementations and low-cost access to robotic hardware. We evaluate submissions in simulation online, on standardized hardware environments, and finally at the competition event. The first AI-DO, AI-DO 1, occurred at the Neural Information Processing Systems (NeurIPS) conference in December 2018.
The results of AI-DO 1 highlight the need for better benchmarks, which are lacking in robotics, as well as improved mechanisms to bridge the gap between simulation and reality.
}


% Sections
% ------------------------------------------
\section{Introduction}

Competitions provide an effective means to advance robotics research by making solutions comparable and results reproducible~\cite{behnke2006robot} as without a clear benchmark of performance, progress is difficult to measure. 
The overwhelming majority of benchmarks and competitions in \ac{ML} do not involve physical robots.
Yet, the interactive embodied setting is thought to be an essential scenario to study intelligence~\cite{pfeifer2001understanding, floreano2004evolution}. 
Robots are generally built as a composition of blocks (perception, estimation, planning, control, etc. ), and an understanding of the interaction between these different components is necessary to make robots achieve tasks. 
However, it is currently unclear whether these abstractions are necessary within \ac{ML} methods, or whether it is possible to learn them in an end-to-end fashion.

In order to evaluate \ac{ML} solutions on physically embodied systems, we developed a competition using the  robotics platform Duckietown - a miniature town on which autonomous robots navigate~\cite{paull2017duckietown}. 
This competition is aimed as a stepping stone to understand the role of AI in robotics in general, and in self-driving cars in particular. 
We call this competition the ``AI Driving Olympics'' (AI-DO) because it comprises a set of different trials that correspond to progressively sophisticated vehicle behaviors, from the reactive task of lane-following to more complex and ``cognitive'' behaviors, such as dynamic obstacle avoidance, to finally coordinating a vehicle fleet while adhering to a set of ``rules of the road.''
The first AI-DO event (AI-DO 1) took place as a live competition at the 2018 Neural Information Processing Systems (NeurIPS) conference in Montr\'eal, Canada. AI-DO 2 will take place at the 2019 International Conference on Robotics and Automation (ICRA), again in Montr\'eal, Canada. 

\begin{figure}[t]
\centering
\includegraphics[width=0.7\textwidth]{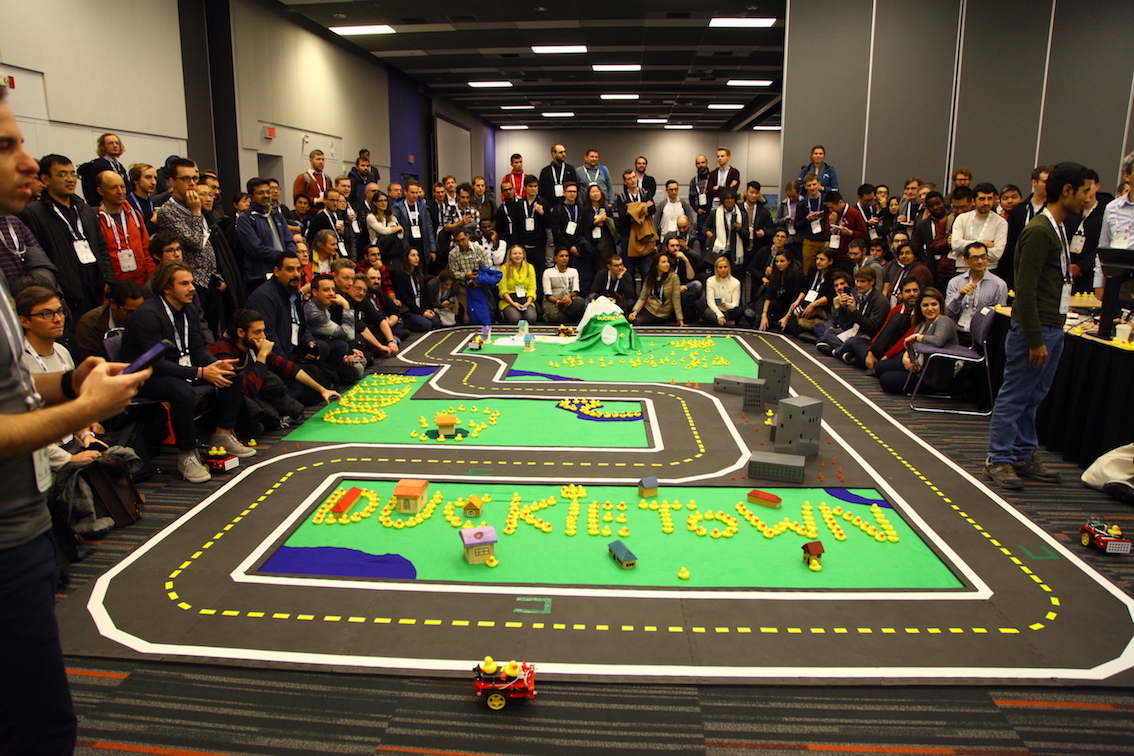}
\caption{The AI-DO 1 competition at NeurIPS 2018, Montr\'eal, Canada.}
\label{fig:competition_overview}
\end{figure}

Here, we describe our overall design objectives for the AI-DO (Sec.~\ref{sec:purpose}). Subsequently, we detail AI-DO 1, describe the approaches taken by the top competitors (Sec.~\ref{sec:results}) and then conduct a critical analysis of which aspects of AI-DO 1 worked well and which did not (Sec.~\ref{sec:lessons-learned}). We then focus our attention on the planned improvements for AI-DO 2 (Sec.~\ref{sec:icra}) and provide some general conclusions (Sec.~\ref{sec:conclusion}).
% ------------------------------------------
\section{The Purpose of AI-DO}\label{sec:vision}
\label{sec:purpose}

\begin{table}[tb]

\centering
	 {\small
	 \begin{tabularx}{0.875\linewidth}{lcccccc}
	   \toprule
	    &
	   \textbf{DARPA~\cite{darpa_grand_challenge}}  &
	   \textbf{KITTI~\cite{kitti}}  &
	   \textbf{Robocup~\cite{robocup}} &
       \textbf{RL comp.~\cite{learning_to_run}} & 
	   \textbf{HuroCup~\cite{hurocup}} &
       \textbf{AI-DO}\\
	   \midrule
	   Accessibility & - & \checkmark &  \checkmark & \checkmark & \checkmark & \bf{\checkmark}   \\
	   Resource constraints & \checkmark & - &  \checkmark  & - & \checkmark & \bf{$\circlearrowright$}  \\
	   Modularity & - & - &  - & - & -  & $\circlearrowright$ \\
       Simulation/Realism & \checkmark & \checkmark & \checkmark & - & \checkmark & \bf{$\circlearrowright$} \\
       \ac{ML} compatibility & - & \checkmark & - &  \checkmark & - & \bf{\checkmark} \\
       Embodiment & \checkmark & - & \checkmark & - & \checkmark & \bf{\checkmark} \\
       Diverse metrics & \checkmark & - &  \checkmark & - & \checkmark & \bf{$\infty$}  \\
       Reproducible experiments & - & \checkmark & - & \checkmark & - & \bf{\checkmark} \\
       Automatic experiments & - & \checkmark & - & \checkmark & - & \bf{$\circlearrowright$} \\
	   \bottomrule
	 \end{tabularx}}
	 \caption{\textbf{Characteristics of various robotic competitions and how they compare to AI-DO} \newline
	 Definitions of characteristics as they pertain to AI-DO are available in Table~\ref{tab:considerations}. 
	 A \checkmark signifies that a competition currently possesses a characteristic. Characteristics with  $\circlearrowright$ are in development or to be improved for AI-DO 2. 
	 Finally, $\infty$ symbolizes features to be added in later editions of AI-DO. With every iteration of the competition, we aim to increase the complexity of the challenges to incorporate what we believe are key characteristics of an ML-based robotics competition.}
	 \label{tab:competition_characteristics}
\end{table}
Robotics is often cited as a potential target application in \ac{ML} literature (e.g., \cite{Singh, DARLA} and many others). The vast majority of these works evaluate their results in simulation~\cite{learning_to_run} and on very simple (usually grid-like) environments~\cite{sutton2018reinforcement}. However, simulations are by design based on what we already know and lack some of the surprises the real world has to offer. 
Our experience thus far indicates that many of the implicit assumptions made in the \ac{ML} community may not be valid for real-time physically embodied systems. For instance, considerations related to resource consumption,
latency, and system engineering are rarely considered but are crucially important for fielding real robots.
Likewise, relying on a single metric to judge behavior (cf. reinforcement learning) is not realistic in more complex tasks. 
Training agents in simulation has become very popular, but it is largely unknown how to assess whether knowledge learned in simulation will transfer to a real robot.  

To some extent, \ac{ML} practitioners can be forgiven for not testing their algorithms on real robot hardware since it is time-consuming and often requires specific expertise.  

Furthermore, running experiments in an embodied setting generally limits: the amount of data that can be gathered, the ease of reproducing previous results, and the level of automation that can be expected of the experiments. To this end, we designed AI-DO as a benchmark that is both trainable efficiently 
in simulation and testable on standardized robot hardware without any robot-specific knowledge.   

What makes the AI-DO unique is that, of the many existing competitions in the field of robotics, none possess the essential characteristics that help facilitate the development of learning from data to deployment. A comparative analysis of these characteristics is given in  
Table~\ref{tab:competition_characteristics} with the details  for AI-DO given in Table~\ref{tab:considerations}.

Ideally, 
the \ac{ML} community would redirect its efforts 
towards physical agents acting in the real world and help elucidate
the unique characteristics of embodied learning in the context of robotics. 
Likewise, 
the robotics community should devote more effort to the use of \ac{ML} techniques where applicable. 
The guaranteed impact is the establishment of a standardized baseline for comparison of learning-based and classical robotics approaches.  

\begin{table}[tb]
\centering
	 \begin{tabular}{|m{0.2\columnwidth}|>{\arraybackslash}p{0.8\columnwidth}|}
	 \toprule
	 \textbf{Characteristic}  & \textbf{Description and Implementation in AI-DO} \\ \hline
	 Accessibility   &  No up front costs other than the option of assembling a ``Duckiebot'' are required. The Duckiebot comes with step-by-step instructions including online tutorials, a forum, and detailed descriptions within the open-source Duckiebook~\cite{Duckiebook}. \\ \hline
    Resource constraints & Constraints on power, computation, memory, and actuator limits play a vital role within robotics.
In particular, we seek to compare contestants with access to different computational substrates such as a Raspberry PI 3 or a Movidius stick with more or less computational speed. 
Following this agenda, we want to explore which method is best given access to a fast computer, which is best using a slower computer, or which is best with more or less memory, etc.   \\ \hline
    Modularity &  More often than not, robotics pipelines can be decomposed into several modules. The optimal division of a pipeline into modules however is undecided. In future editions of AI-DO we will make such modules explicit and test them according to performance and resource consumption such as computational load and memory.  \\ \hline
    Simulation/realism &  The AI-DO is an ideal setup to test sim-to-real transfer wince both simulation and real environment are readily available for testing and the software infrastructure enables toggling between the two. \\ \hline
    ML compatibility &  Logged data from hours of Duckiebot operations are available to allow for training of ML algorithms. We also provide an  OpenAI Gym~\cite{1606.01540}  interface for both the simulator and real robot, which enables easy development of reinforcement learning algorithms.   \\ \hline
    Embodiment &  The AI-DO finals are run in a real robot environment (Duckietown).  \\ \hline
    Diverse metrics &   Each challenge of AI-DO employs specific objective metrics to quantify success, such as:
\textit{traffic law compliance metrics} to penalize illegal behaviors (e.g., not respecting the right-of-way);
\textit{performance metrics} such as the average speed, to penalize inefficient solutions and
\textit{comfort metrics} to penalize unnatural solutions that would not be comfortable to a passenger.
These metrics are based on measurable quantities such as speed, timing and detecting traffic violations. 
Consequently,  there may be multiple winners in each competition (e.g., a very conservative solution, a more adventurous one, etc.).   \\ \hline
    
Reproducible \qquad \qquad experiments & The overall objective is to design experiments that can easily be reproduced, similar to recent trends in RL \cite{henderson2018deep}.
We facilitate reproducibility by running available code on standard hardware and containerization of software. \\ \hline
Automatic experiments &  
Embodiment makes the problem of standardized evaluations particularly challenging. The ability to automatically conduct controlled experiments and perform evaluation will help to address these challenges. To that end,
we are building intelligent Duckietowns called \emph{Robotariums}~\cite{pickem2017robotarium}  to automate the deployment and evaluation of autonomy algorithms on standardized hardware. 
\\ \hline
\end{tabular}
\caption{Important characteristics for learning-based robotics competitions}
\label{tab:considerations}
\end{table}

% ------------------------------------------
\section{AI-DO 1: Competition Description}\label{sec:competition}

In the following, we describe the first version of the AI Driving Olympics (AI-DO). In particular we detail:
\begin{itemize}
    \item The different competition challenges and their associated scoring metrics;
    \item The physical Duckietown platform; 
    \item The software infrastructure including simulators and logs, the Docker containerization infrastructure, and various baseline implementations.
\end{itemize}

\subsection{The Challenges}
\begin{figure}[!t]
    \centering
	\vspace{2mm}
    \includegraphics[height=3.1cm]{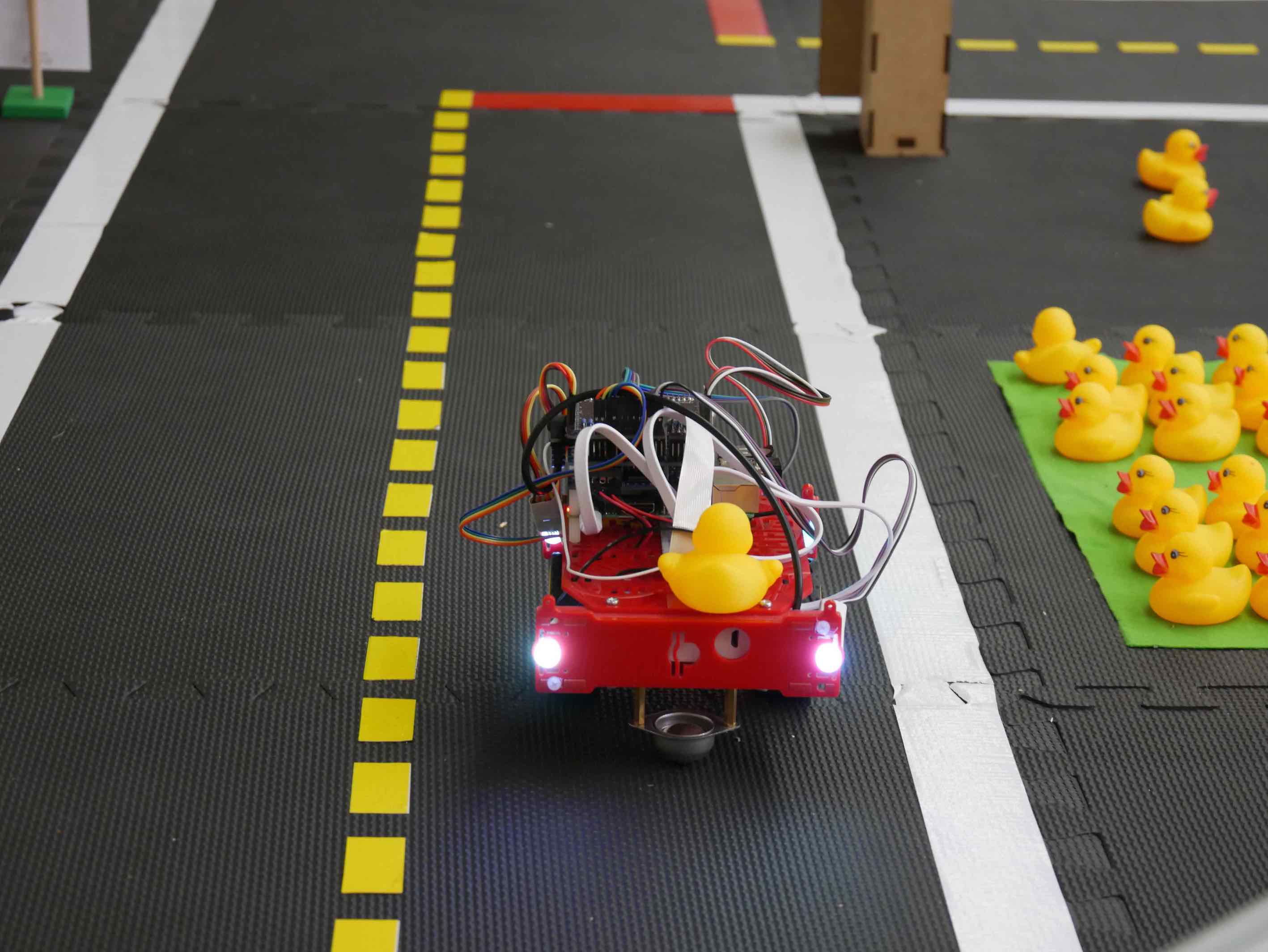} \hfil%
    \includegraphics[height=3.1cm]{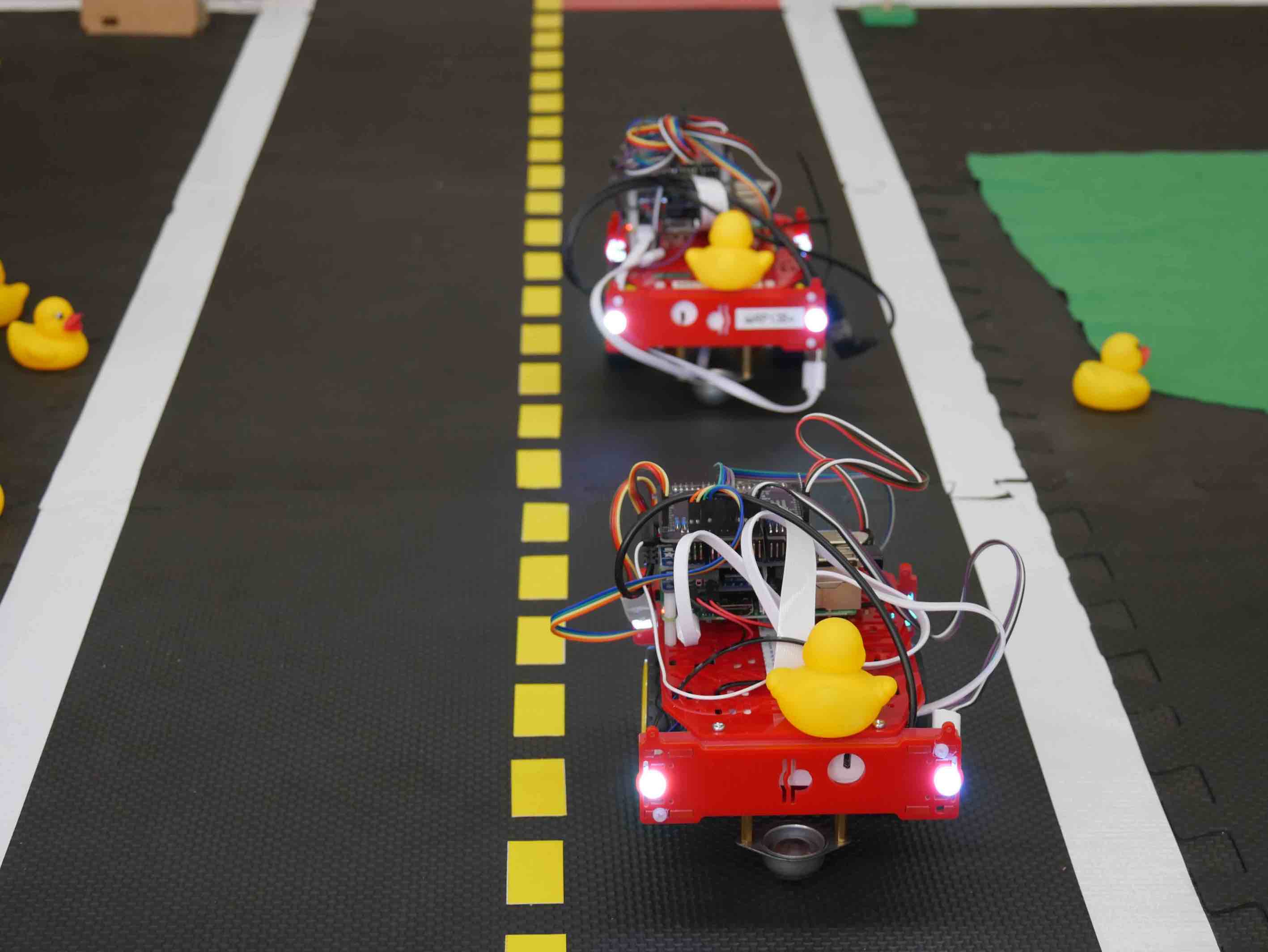} \hfil%
    \includegraphics[height=3.1cm]{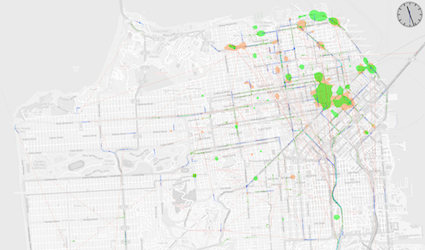}
	\caption{The tasks in AI-DO 1 included (left) lane following, (middle) lane following with obstacles, and (right) autonomous mobility on demand.}
	\label{tab:task_imgs}
\end{figure}
AI-DO 1 comprised three challenges with increasing order of difficulty:
\begin{enumerate}
    \item[\footnotesize{\textbf{LF}}] \textbf{Lane following} on a continuous closed course, \textbf{without obstacles}. The robot was placed on a conforming closed track (with no intersections) and was required to follow the track in the right-hand lane. 
    \item[\footnotesize{\textbf{LFV}}] \textbf{Lane following} on a continuous closed course as above, but 
	\textbf{with static obstacles (e.g., vehicles)} sharing the road. 
    \item[\footnotesize{\textbf{AMOD}}] \textbf{Autonomous mobility on demand}:\footnote{AMOD competition website \url{https://www.amodeus.science/}} Participants were required to implement a centralized dispatcher that provided goals to individual robots in a fleet of autonomous vehicles in order to best serve customers requesting rides. Low-level control of the individual vehicles was assumed. 
\end{enumerate}

\subsection{Metrics and Judging}

The rules for AI-DO 1 are described below and can be found in more detail online.\footnote{The performance rules of AI-DO \url{http://docs.duckietown.org/DT18/AIDO/out/aido_rules.html}} 

\subsubsection{Lane following with and without obstacles (LF/LFV)}
While not treated separately in AI-DO 1 for the lane following (LF) and lane following with obstacles (LFV) task, we considered the following metrics, each evaluated over five runs to reduce variance.

\textbf{Traveled distance:}
The median distance traveled along a discretized lane. Going in circles or traveling in the opposite lane does not affect the distance.

\textbf{Survival time:}
The median duration of an episode, which terminates when the car navigates off the road or collides with an obstacle.

\textbf{Lateral deviation:}
The median lateral deviation from the center line.

\textbf{Major infractions:}
The median of the time spent outside of the driveable zones. For example, this penalizes driving in the wrong lane.

The winner was determined as the user with the longest traveled distance. 
In case of a tie, the above metrics acted as a tie-breaker such that a longer survival time as well as a low lateral deviation and few infractions were rewarded.

\subsubsection{Autonomous mobility on demand (AMOD)}

The autonomous mobility on demand challenge assumed that vehicles were already following the rules-of-the-road and were driving reasonably. 
The focus here was on deciding where and when to send each vehicle to pick up and deliver customers. 
To this end, we used three different metrics to evaluate performance.

\textbf{Service quality:} In this arrangement, we penalized large waiting times for customers as well as large distances driven around with empty vehicles. We weighted the two components such that waiting times dominated the metric. 

\textbf{Efficiency:} Again, we penalized large waiting times for customers as well as large distances driven around with empty vehicles.  In this case, we weighted the two components such that the distance traveled by empty vehicles dominated the metric. 

\textbf{Fleet size:} 
This scenario considered the true/false case of whether a certain service level could be upheld with a chosen fleet size or not. The smaller the fleet the better, however only as long as service requirements were fulfilled.

\subsection{The Physical Competition Infrastructure}\label{sec:platform-physical}

The physical Duckietown platform is comprised of intelligent vehicles (\textit{Duckiebots}) and a customizable model of urban environment (\textit{Duckietown}).

\textbf{The robot - Duckiebot:}
\begin{figure}[!t]
    \centering
    \includegraphics[width=0.5\textwidth]{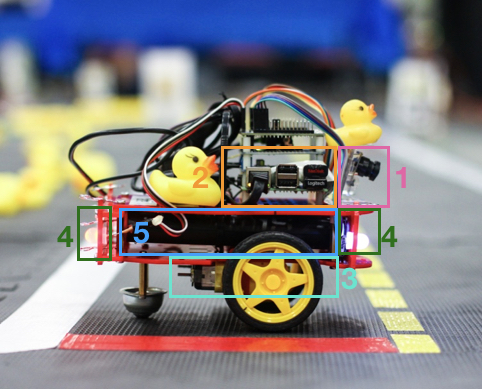}
    \caption{The main components of a Duckiebot. Numbered boxes correspond to components listed in Tab.~\ref{tab:db-components}.}
    \label{fig:db-components}
\end{figure}
Table~\ref{tab:db-components} lists the main components of the Duckiebot robot, which are visualized in Fig.~\ref{fig:db-components}.
\begin{table*}[ht]
    \caption{\emph{Main components of a Duckiebot.}}
    \centering
    \begin{tabularx}{0.9\linewidth}{clll}
    \toprule
    Number  &  Task &  Component & Description\\
    \midrule
    1   &   Sensing    & Fish-eye Camera      & $160^{\circ}$ FOV (front-facing), $640 \times 420$ @30Hz\\
    2   &   Computation   & Raspberry Pi 3B+    & Quad Core 1.4 GHz, 64 bit CPU and 1 GB of RAM\\
    3   &   Actuation     & $2\times$ DC Motors & Independent, in differential drive configuration\\
    4   &   Communication & $5\times$ Addressable LEDs & $3\times$ Front, $2\times$ Back for inter-bot communication (blinking) \\
    5   &   Power & $1\times$ Power Bank & 10000 mAh battery ($>5$h operation) \\         
    \bottomrule
    \end{tabularx}
    \label{tab:db-components}
\end{table*}
We carefully chose these components in order to provide a robot capable of advanced single- and multi-agent behaviors, while maintaining an accessible cost. We refer the reader to the Duckiebook~\cite{Duckiebook} for further information about the robot hardware.

\textbf{The environment - Duckietown:} 

\begin{figure}[t]
    \centering
    \includegraphics[width=0.6\textwidth]{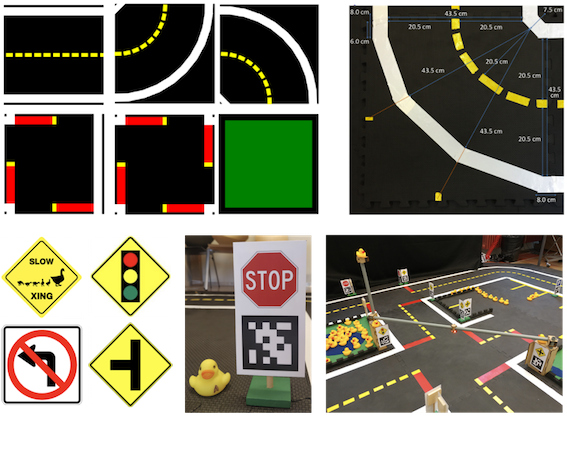}
    \caption{The Duckietown environment is rigorously defined at the road and signal level. When the appearance specifications are met, Duckiebots are guaranteed to navigate cities of any conforming topology.}
    \label{fig:duckietown-environment}
\end{figure}

Duckietowns are modular, structured environments built on \textit{road} and \textit{signal} layers (Fig.~\ref{fig:duckietown-environment}) that are designed to ensure a reproducible driving experience. 

A town consists of six well defined \textit{road segments}: straight, left and right $90^\circ{}$ turns, a 3-way intersection, a 4-way intersection, and an empty tile. Each segment is built on individual interlocking tiles, that can be reconfigured to customize the size and topology of a city. Specifications govern the color and size of the lane markings as well as the road geometry.

The \textit{signal layer} is comprised of street signs and traffic lights. 
In the baseline implementation, street signs are paired with AprilTags~\cite{AprilTags} to facilitate global localization and interpretation of intersection topologies by Duckiebots. 
The appearance specifications detail their size, height and location in the city. 
Traffic lights provide a centralized solution for intersection coordination, encoding signals through different frequencies of blinking LED lights.

These specifications are meant to allow for a reliable, reproducible experimental setup that would be less repeatable on regular streets. For more details about the components and the rules of a Duckietown, we refer the reader to the Duckietown operation manual~\cite{Duckiebook}.

% ------------------------------------------
%%%%%%%%%%%%%%%%%%%%%%%%%%%%%%%%%%%%%%%%%%%%%%%%%%%%%%%%%%%%%%%%%%
% - what runs on the bot (docker)
% - how the simulators use the same stack -> portability
\subsection{The Software Infrastructure}\label{sec:platform-software}

\begin{figure}[!t]
    \centering
    \includegraphics[width=0.9\textwidth]{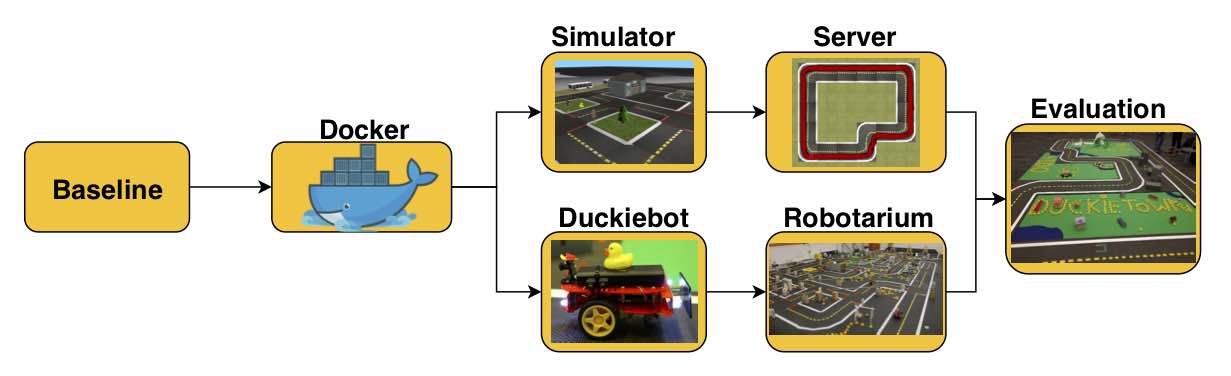}
    \caption{Illustration showing the interconnected components parts of AI-DO 1 evaluation architecture, starting from existing baseline implementations, to Docker-based deployments in simulation and on real Duckiebots, to automatic evaluations of driving behavior.}
    \label{fig:software_overview}
\end{figure}
%
%
% 1. Philosophy of software design - the design principles
The overarching design principles underlying our software architecture are the following:
\begin{itemize}
    \item Minimize dependencies on specific hardware or operating systems for development, which we achieved via containerization with Docker;
    \item Standardize interfaces for ease of use, specifically to allow a participant to easily switch between developing driving solutions in simulation and deploying them on physical robots (either their own or in a Robotarium). This same interface is used to evaluate submissions;
    \item Provide interfaces to commonly used frameworks in both the ML (e.g., Tensorflow, PyTorch, and OpenAI Gym) and robotics (e.g., ROS) communities;
    \item Provide baseline implementations of algorithms commonly used in both ML (e.g., reinforcement learning, imitation learning) and robotics (see \citet{paull2017duckietown} for details) communities;
    \item Provide tools to leverage cloud resources such as GPUs.
\end{itemize}
We use Docker containerization to standardize the components as well as inter-component communication.
This has the important advantage that it allows us to deploy the same code in the simulator as well as on the real Duckiebot using an 
identical interface.
Sec.~\ref{sec:docker} contains more details about our different container types and the interactions between them, while Fig.~\ref{fig:software_overview} depicts the evaluation architecture.

% 2. What components there are and what they are useful for
As mentioned above, in order to facilitate fast training with modern machine learning algorithms, we developed an OpenGL-based simulator ``gym-duckietown''~\cite{gym_duckietown}. As the name suggests, the simulator provides an OpenAI gym-compatible interface~\cite{1606.01540} that enables the use of available implementations of state-of-the-art reinforcement learning algorithms. Sec.~\ref{sec:simulation} discusses the simulation and its features in more detail.

Our architecture is unique in that it requires that all submissions be containerized. Rather than directly connecting to the simulator, each participant launches and connects to a Docker container that runs the simulation. In order to facilitate this interaction, we provide  boilerplate code for making submissions and for launching the simulator. We also provide baseline autonomous driving implementations based upon imitation learning, reinforcement learning, as well as a classic ROS-based lane detection pipeline. Each of these three implementations provides basic driving functionality out-of-the-box, but also includes pointers to various improvements to the source code consistent with the goal that AI-DO be educational. 

% 3. The user experience
Once a user submits an agent container, our cluster of evaluation servers downloads and scores the submission which is elaborated in Sec.~\ref{sec:dev-pipeline}. During this process, the servers create various forms of feedback for the user. These include various numerical evaluation scores as discussed earlier (e.g., survival time, traffic infractions, etc.), plots of several runs overlaid on a map, animated GIFs that show the robot's view at any given time, and log files that record the state of the system during these runs. The evaluation server uploads this data to a leaderboard server.

% 4. How software runs in the competition, submitting and evaluating the performance
We provided the code and documentation for running the submissions on the physical Duckiebot. Originally, we intended for this to happen autonomously in the Robotarium.
Since this was not completed before the first AI-DO live challenge, we instead ``manually'' ran the containers associated with the best simulation results on the physical systems. 
Specifically, we ran a Docker container on our Duckiebots that 
provided low-level control and perception with low latency while
exposing to the local WiFi network the same ROS interface 
provided in the simulator.
From a PC connected to the same local network, we downloaded the agent
submissions (Docker images) and run it so that the agent container
was able to control the Duckiebot over the network.

\subsubsection{The Development Pipeline}\label{sec:dev-pipeline}

The AI-DO development process was designed with two primary goals in mind. First, it should be easy for competitors to install and run prepared ``baselines'' (Sec.~\ref{subsec:baselines}) for each of the challenges. Secondly, it should be easy to test and evaluate each submission in a controlled and reproducible manner. To do so, we implemented a command line utility for building and running solutions, and a set of containerization tools (Sec.~\ref{sec:docker}), enabling users to quickly fetch the necessary dependencies and evaluate their submissions locally.

For lane-following, we provided the following baseline implementations: Classic ROS, reinforcement learning (RL) with PyTorch, and imitation learning (IL) with Tensorflow. All provided baselines ran unmodified in the simulator, and some (e.g., IL) were trained using logs recorded from human drivers. We also provided docker images for producing trained RL and IL models. Following internal testing and evaluation, we released baselines to the public during the competition period.

\begin{figure}[!th]
    \centering
    \includegraphics[width=0.6\textwidth]{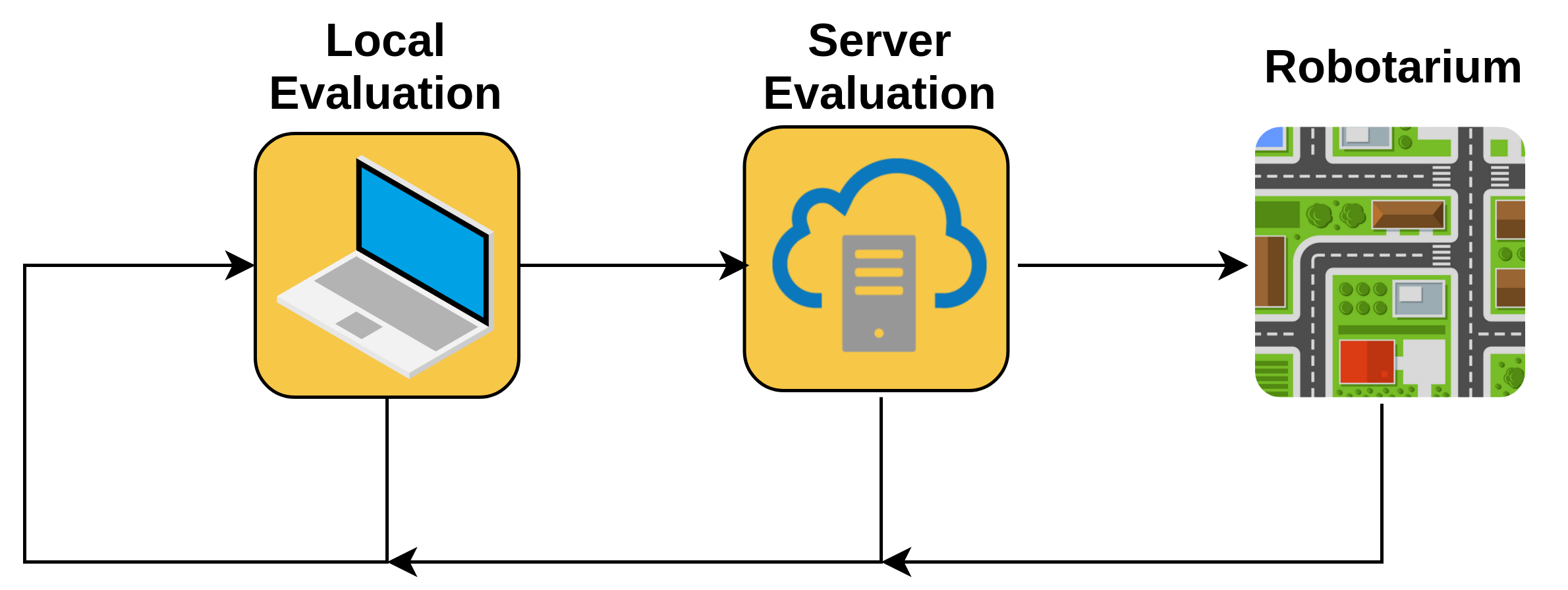}
    \caption{AI-DO submissions must pass a series of checkpoints in order to qualify for the evaluation phase.} 
    \label{fig:evaluation-checkpoints}
\end{figure}
We encouraged competitors to use the following protocol for submission of their challenge solutions (Fig.~\ref{fig:evaluation-checkpoints}). First, they should build and evaluate a baseline inside the simulator on their local development machine to ensure it was working properly. This evaluation results in a performance score and data logs that are useful for debugging purposes. If the resulting trajectory appeared satisfactory, competitors could then submit the Docker image to the official AI-DO evaluation server. 
The server assigned all valid submissions a numerical score and placed them on a public leaderboard.

While the submission of competition code via containerization was straightforward, one issue we encountered during internal development was related to updates to the Docker-based infrastructure (see Sec.~\ref{sec:docker}).
Originally, Docker images would be automatically rebuilt upon (1) changes to the mainline development branch and (2) changes to base layers, resulting in a cascade of changes to downstream Docker images. This was primarily due to two issues: (1) images were not version-pinned and (2) the lack of acceptance testing. Since users were automatically updated to the latest release, this would cause a significant disruption to the development workflow, where cascading failures were observed on a daily basis. To address this issue, we disabled auto builds and deployed manually, however a more complete solution would involve rolling out builds incrementally following a standard acceptance testing process. Due to a premature automation of the build pipeline, we were unable to utilize auto-builds to their full capabilities.

\subsubsection{Simulation}\label{sec:simulation}

\begin{figure}[!th]
    \centering
    \includegraphics[height=3.3cm]{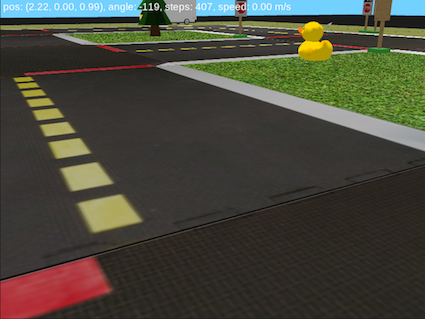}\hfil
    \includegraphics[height=3.3cm]{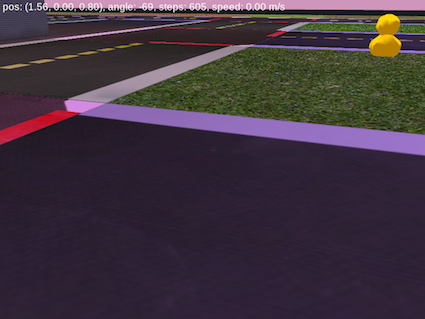}\hfil
    \includegraphics[height=3.3cm]{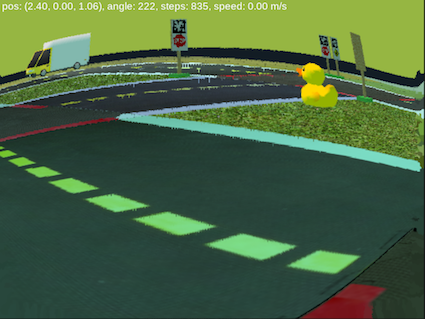}
    \caption{The lightweight OpenGL-based simulator provides (left) realistic synthetic images, (middle) visual domain randomization, and (right) lens distortion effects.}
\label{fig:simulator}
\end{figure}

While Duckietown's software architecture allows for fast deployment on real robots, it is often easier and more efficient to test new ideas within simulated environments. Fast simulators are also important when developing and training reinforcement learning or imitation learning algorithms, which often require large amounts of data. Duckietown ships with a fast OpenGL-based simulator~\cite{gym_duckietown} (Fig.~\ref{fig:simulator}) that incorporates realistic dynamic models and simulates various Duckietown maps within a purely Python framework.  The simulator runs at hundreds of frames per second, and is fully customizable in all aspects that control it, including dynamics, environment, and visual fidelity.

The simulator provides several features that help model operation in the real Duckietown environment. These include the simulation of other agents including Duckiebots and pedestrian Duckies, as well as many of the obstacles found in physical Duckietowns, such as traffic cones and cement dividers. The simulator ships with multiple maps and a random map generator, enabling an easy way for participants to ensure that their agents do not overfit to various details of the environment. The simulator provides high-fidelity simulated camera images that model realistic fish-eye lens warping.

However, it is often the case that algorithms trained purely in simulation fall prey to the \textit{simulation-reality gap}~\cite{jakobi1995noise} and fail to transfer to the real Duckiebot. This gap is particularly problematic for methods that reason over visual observations, which often differ significantly between simulation and the real environment. An increasingly common means of bridging the simulation-to-reality gap is to employ domain randomization~\cite{Tobin2017DomainWorld}, which randomizes various aspects of the simulator, such as colors, lighting, action frequency, and various physical and dynamical parameters, in the hope of preventing learned policies from overfitting to simulation-specific details that will differ after transfer to reality. 
The Duckietown simulator provides the user with hooks to dozens of parameters that control domain randomization, all with configurable ranges and settings through a JSON API. 

Simulators are integral to the development process, particularly for learning-based methods that require access to large datasets. In order to ensure fairness in the development process, we also provide wrappers that allow traditional (non-learning-based) robotic architectures to benefit from the use of simulation. Participants are encouraged to try combinations of various approaches, and are able to quickly run any embodied task such as lane following (LF) and lane following with obstacles (LFV) within the simulator by changing only a few arguments. For the autonomous mobility on demand (AMOD) task, we provide a separate city-based fleet-level simulator~\cite{ruch2018amodeus}.

\subsubsection{Containerization}\label{sec:docker}

One of the challenges of distributed software development across heterogeneous platforms is the problem of variability. With the increasing pace of software development comes the added burden of software maintenance. As hardware and software stacks evolve, so too must source code be updated to build and run correctly. Maintaining a stable and well documented codebase can be a considerable challenge, especially in a robotics setting where contributors are frequently joining and leaving the project. Together, these challenges present significant obstacles to experimental reproducibility and scientific collaboration.

In order to address the issue of software reproducibility, we developed a set of tools and development workflows that draw on best practices in software engineering. These tools are primarily built around containerization, a widely adopted virtualization technology in the software industry. In order to lower the barrier of entry for participants and minimize variability across platforms (e.g. simulators, robotariums, Duckiebots), we provide a state-of-the-art container infrastructure based on Docker, a popular container engine. Docker allows us to construct versioned deployment artifacts that represent the entire filesystem and to manage resource constraints via a sandboxed runtime environment.

\begin{figure}[ht]
    \centering
    \includegraphics[width=0.45\textwidth]{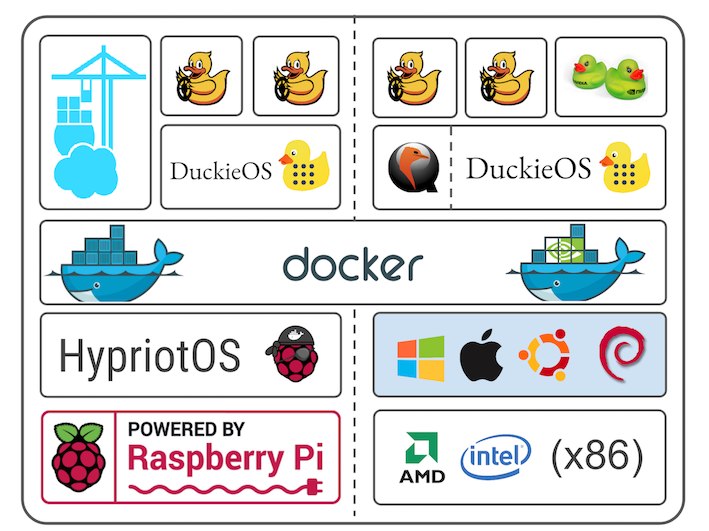} \hfil
    \includegraphics[width=0.45\textwidth]{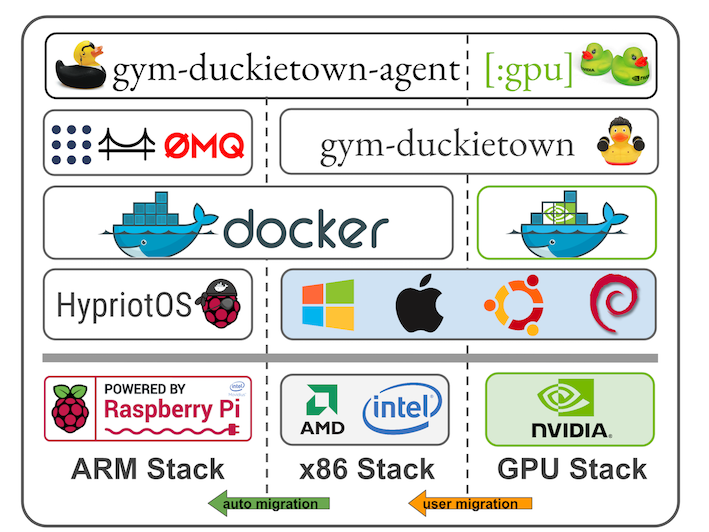}
    \caption{AI-DO container infrastructure. Left: The ROS stack targets two primary architectures, x86 and ARM. To simplify the build process, we only build ARM artifacts, and emulate ARM on x86. Right: Reinforcement learning stack. Build artifacts are typically trained on a GPU, and transferred to CPU for evaluation. Deep learning models, depending on their specific architecture, may be run on an ARM device using an Intel NCS.}
\label{fig:docker}
\end{figure}

The Duckietown platform supports two primary instruction set architectures: x86 and ARM. To ensure the runtime compatibility of Duckietown packages, we cross-build using hardware virtualization to ensure build artifacts can be run on all target architectures. Runtime emulation of foreign artifacts is also possible, using the same technique.\footnote{For more information, this technique is described in further depth at the following URL: \url{https://www.balena.io/blog/building-arm-containers-on-any-x86-machine-even-dockerhub/}.} For performance and simplicity, we only build ARM artifacts and use emulation where necessary (e.g., on x86 devices). On ARM-native, the base operating system is HypriotOS, a lightweight Debian distribution with built-in support for Docker. For both x86 and ARM-native, Docker is the underlying container platform upon which all user applications are run, inside a container.

Docker containers are sandboxed runtime environments that are portable, reproducible and version controlled. Each environment contains all the software dependencies necessary to run the packaged application(s), but remains isolated from the host OS and file system. Docker provides a mechanism to control the resources each container is permitted to access, and a separate Linux namespace for each container, isolating the network, users, and file system mounts. Unlike virtual machines, container-based virtualization like Docker only requires a lightweight kernel, and can support running many simultaneous containers with close to zero overhead. A single Raspberry Pi is capable of supporting hundreds of running containers.

While containerization considerably simplifies the process of building and deploying applications, it also introduces some additional complexity to the software development lifecycle. Docker, like most container platforms, uses a layered filesystem. This enables Docker to take an existing ``image'' and change it by installing new dependencies or modifying its functionality. Images may be based on a number of lower layers, which must periodically be updated. Care must be taken when designing the development pipeline to ensure that such updates do not silently break a subsequent layer as described earlier in Sec.~\ref{sec:dev-pipeline}.

One issue encountered is the matter of whether to package source code directly inside the container, or to store it separately. If source code is stored separately, a developer can use a shared volume on the host OS for build purposes. In this case, while submissions may be reproducible, they are not easily modified or inspected. The second method is to ship code directly inside the container, where any changes to the source code will trigger a subsequent rebuild, effectively tying the sources and the build artifacts together. Including source code alongside build artifacts also has the benefit of reproducibility and diagnostics. If a competitor requires assistance, troubleshooting becomes much easier when source code is directly accessible. However doing so adds some friction during development, which has caused competitors to struggle with environment setup. One solution is to store all sources on the local development environment and rebuild the Docker environment periodically, copying sources into the image.

\subsubsection{Baselines}\label{subsec:baselines}

% \TODO{A LOT CAN BE SAID HERE}
A important aspect of AI-DO is that participants should be able to quickly join the competition, without the significant up-front cost that comes with implementing methods from scratch. To facilitate this, we provided baseline ``strawman'' solutions, based both on ``classical" and learning-based approaches. 
These solutions are fully functional (i.e., they will get a score if submitted) and contain all the necessary components, but should be easily beatable. 

\begin{figure}[t]
    \centering
    \includegraphics[height=3.3cm]{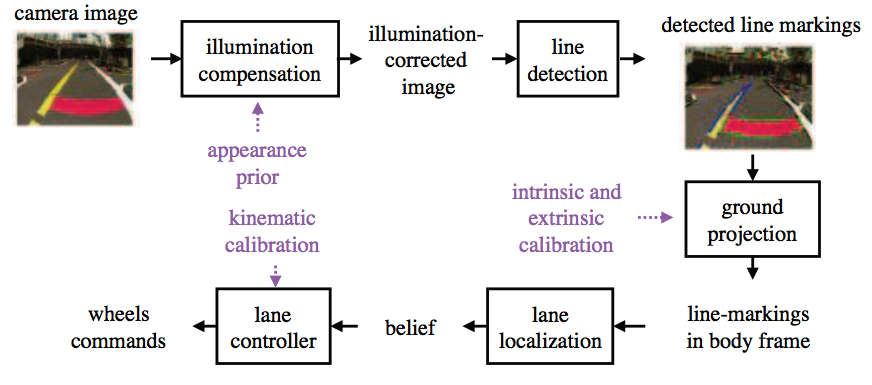} \hfil
    \includegraphics[height=3.3cm]{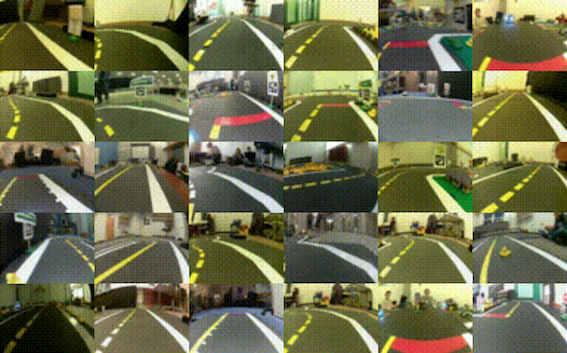}
    \caption{We provided several baselines including (left) a ROS-based Duckietown lane detection and tracking pipeline, and (right) an imitation learning baseline trained on 17 hours of logged data.}
    \label{fig:logs}
\end{figure}

\textbf{ROS baseline:}
We provided a simplified implementation of the classical Duckietown ROS-based baseline (Fig.~\ref{fig:logs} (left)), which consists of a simple lane detection algorithm that maps to wheel velocities. 
It is provided in order to show participants how to quickly use combinations of existing Duckietown software with their own enhancements. 
The ability to mix classical methods into a software pipeline enables participants to overcome various issues that often plague purely learning-based methods. 
For example, when training an end-to-end reinforcement learning policy in simulation, the policy may overfit to various visual simulation details, which upon transfer, will no longer be present. In contrast, running a learning algorithm on the top of the output of a classical module, such as a lane filter, will transfer more smoothly to the real world.

\textbf{Imitation Learning from Simulation or Logs:} 
Learning complex behaviors can be a daunting task if one has to start from scratch. 
Available solutions such as the above ROS-based baseline help to mitigate this. 
As one way to make use of this prior information, we are offering simple imitation learning baselines (behavioral cloning) that utilize on driving behavior datasets collected either in simulation or from recorded logs\footnote{Duckietown logs database: \url{http://logs.duckietown.org/}} as depicted on the right in Fig.~\ref{fig:logs}.
In our experience, policies learned using imitation learning within Duckietown are frequently very skilled at driving as long as sufficient and representative data is selected to train them. 

\textbf{Reinforcement Learning Baseline:}
For participants who want to explore reinforcement learning approaches within AI-DO, we supply a lane following baseline using the Deep Deterministic Policy Gradients (DDPG) method~\cite{lillicrap2015continuous} trained in the Duckietown simulation environment. 
DDPG is an actor-critic method that can deal with continuous output spaces based on the Deep Q-learning architecture~\cite{mnih2015human} originally developed for discrete action spaces.
To aid in improving the baseline, we also provide a list of possible changes that participants can try to improve the baseline. 
The distinguishing factor of such a reinforcement learning solution is that, at least in principle, it has the chance to find unexpected solutions that may not easily be found through engineering insights (ROS baseline) or copying of existing behaviors (imitation learning baseline). 
% ------------------------------------------
%% RESULTS SECTION:
\section{AI-DO 1: Details of Top Submissions} \label{sec:results}

The first edition of the AI Driving Olympics included an initial online phase in which submission were evaluated in simulation. 
We received our first submissions for AI-DO 1 in October 2018. Participants competed for a spot in the final competition up until the day before the final competition on 9 December at NeurIPS 2018 in Montr\'eal, Canada.

\subsection{Qualifications}\label{sec:results-quals}

We received a total of around 2000 submissions from around 110 unique users, with the majority focused on the lane following (LF) task. Only a small number of participants considered the lane following with obstacles (LFV) and autonomous mobility on demand (AMOD) tasks.
As a consequence, we limited the final competition to the lane following task and included the top five participants from the lane following simulation phase. 
These will be described in the following pages. 
We were interested in understanding the different approaches that the participants took and the various design decisions that they made to improve performance.
Afterwards we will share our experiences at the live finals, where we saw the code of the finalists in action on real Duckiebots.  

\begin{figure}
    \centering
    \includegraphics[width=0.3\textwidth, height=4cm]{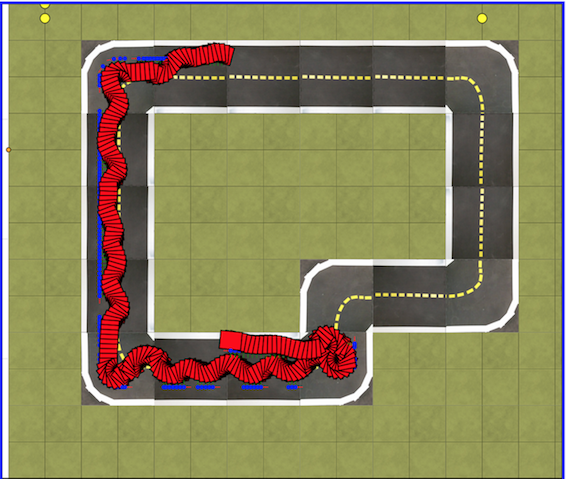}\hfil  %3.35
    \includegraphics[width=0.3\textwidth, height=4cm]{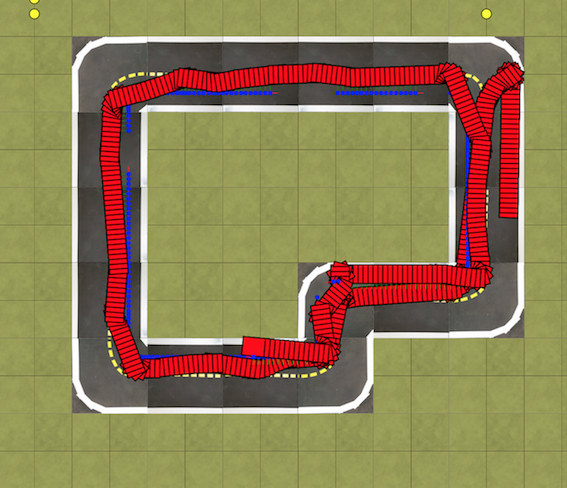}\hfil
    \includegraphics[width=0.3\textwidth, height=4cm]{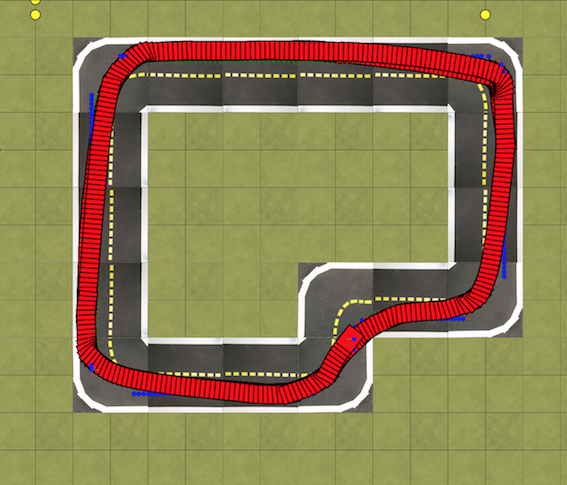} \\
    \includegraphics[width=0.3\textwidth, height=4cm]{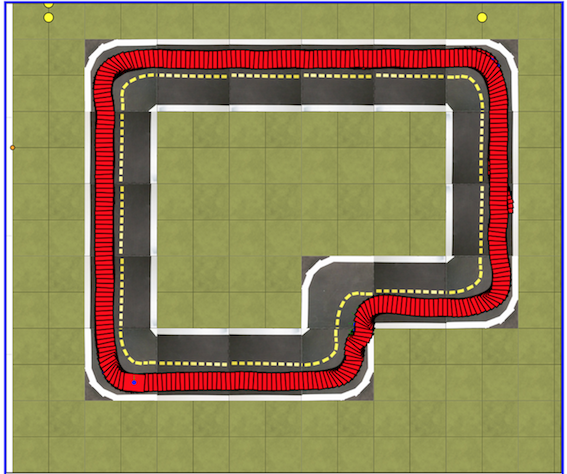}\hfil
    \includegraphics[width=0.3\textwidth, height=4cm]{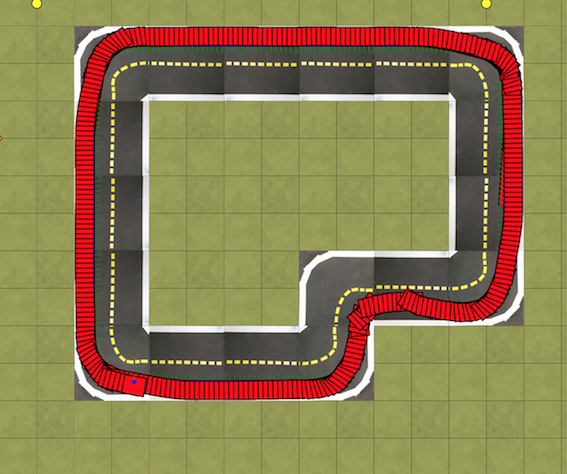}\hfil
    \includegraphics[width=0.3\textwidth, height=4cm]{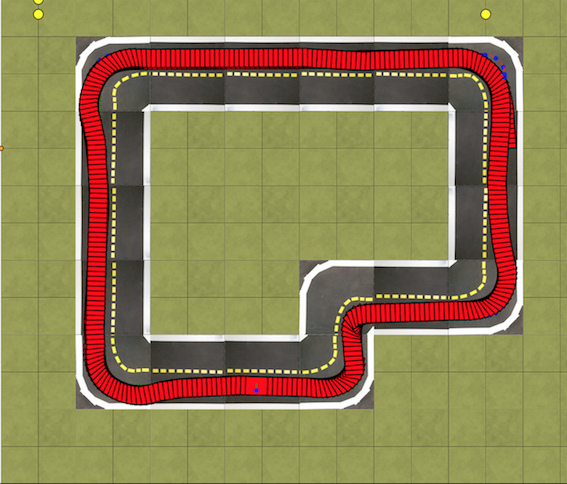}
    \caption{A bird's-eye-view of the trajectories that six different agent submissions took with performance increasing clockwise from the upper-left: a submission that did not reach the finals; that of Vincent Mai; that of Jon Plante; that of Team Jetbrains; that of Team SAIC Moscow; and the winning entry by Wei Gao.}
    \label{fig:driving_behavior}
\end{figure}

To put the competitor submissions into context, we ask the questions of what constitutes bad driving and what good driving is?
As an example of challenges already present when driving in simulation refer to Fig.~\ref{fig:driving_behavior}. 
Depicted is a bird's-eye-view visualization of the driving path of six different submissions. 
We include this figure to demonstrate the spread of behaviors between the submissions of the finalists and another submission (left-top) that did not advance to the finals. 

\clearpage

\subsubsection{Contribution from Participant 1: Gao Wei (Team Panasonic R\&D Center, Singapore \& NUS)}\label{sec:results-cont-1}

\begin{figure}[!t]
    \centering
    \includegraphics[width=0.8\textwidth]{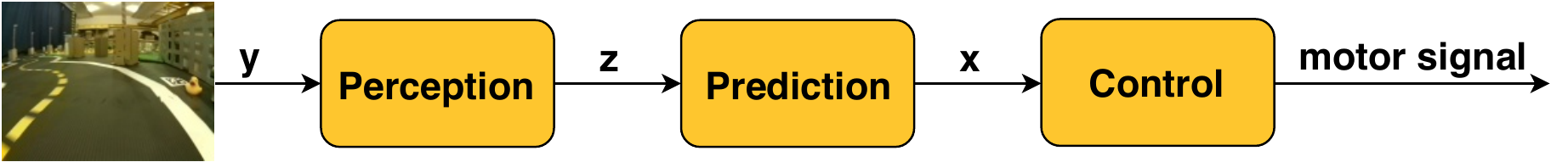}
    \caption{The pipeline of Gao Wei processes an image $y$ to find the lane, its curvature and possible stop lines. 
    These are further transformed to provide a bird's-eye perspective $x$ to the controller.
    The controller then follows a pure-pursuit strategy based on the estimated 3D map.}
    \label{fig:pipeline_gao_wei}
\end{figure}

\textbf{Inspiration:} 
This technique was inspired by human driving. 
A person may reason about the lane and potential curvature of the road, plan a trajectory and then execute the plan. % using throttle and steering wheel. 
This is the approach taken here with the agent first projecting the Duckiebot's first-person view to a local ego-centric reconstruction of the lanes, objects and stop lines. 
Based upon the cumulative observations the method then predicts the curvature of the lane. 
Finally, an adaptive pure pursuit algorithm is run on the reconstructed local world and prediction.

\textbf{Perception:}
The perception module processes images from the forward-facing camera to detect lane, objects and moving cars. 
To simplify the communication pipeline and the difficulty of detection, a semantic mask was designed as the API to transfer to prediction.

\textbf{Prediction:}
The prediction module reconstructs the local 3D environment in real-time from the semantic mask given by the perception module. 
It transfers detections in the first-person view to a bird's-eye view of the environment. The module uses a dynamics model of the Duckiebot to update the environment model as the robot navigates.

\textbf{Control:}
The control module reasons over the predicted environment ``map'' to generate control inputs using a pure pursuit control algorithm.

\begin{figure}[ht]
    \centering
    \includegraphics[width=0.45\textwidth, height=4cm]{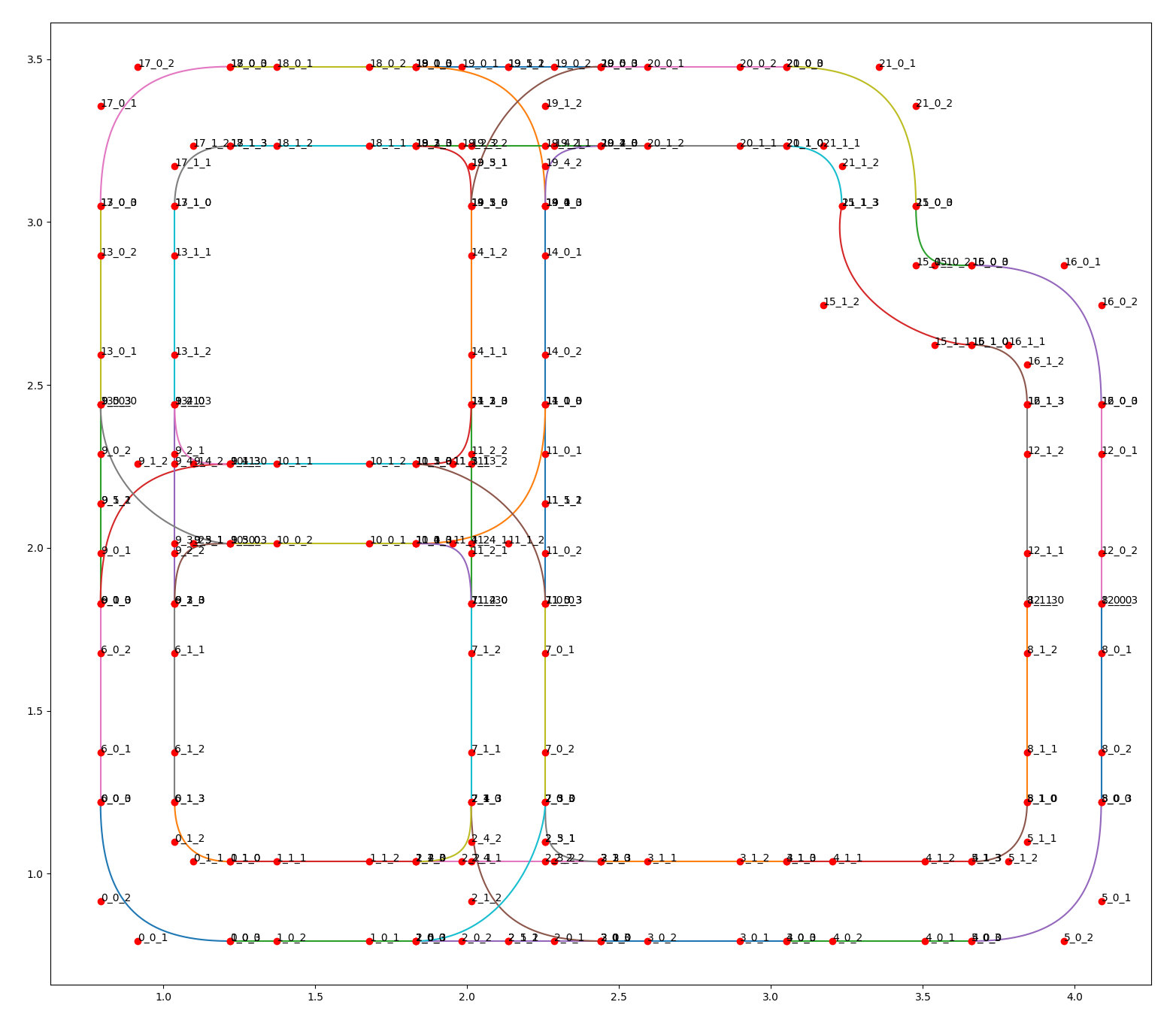} \hfil
    \includegraphics[height=4cm]{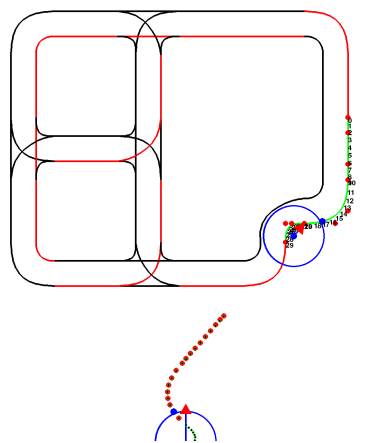}
    \caption{\emph{Left:} A rendering of the ground-truth centerlines from the simulator. \emph{Right:} is a visualization tool for following the idealized center lines.}
    \label{fig:visualization_gao_wei}
\end{figure}
\textbf{Competition experience and lessons:}
In order to facilitate the development process, the participant modified the open-source simulator to access additional information beneficial for debugging, including the ground-truth center line location (Fig.\ref{fig:visualization_gao_wei}). Additionally, they created a custom tool to visualize the inferred egocentric map of the robot's local environment.

\subsubsection{Contribution from Participant 2: Jon Plante}\label{sec:results-cont-2}

\begin{figure}[!t]
    \centering
    \includegraphics[width=0.8\textwidth]{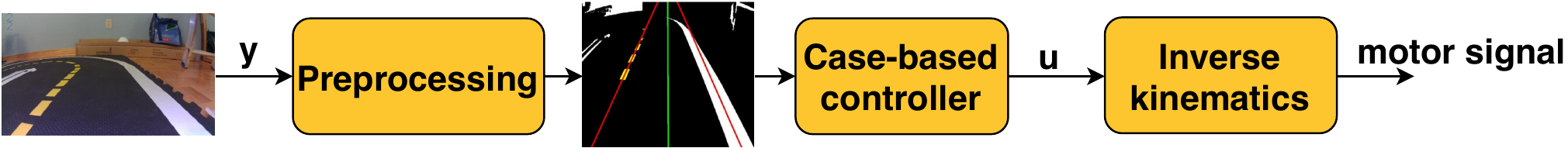}
    \caption{In the approach developed by Jon Plante, images $y$ are preprocessed to remove variability and improve direction estimation to give a position error estimate to a lookup table-based controller. 
    The controller checks in which direction the vehicle intends to go, and then sends the corresponding predetermined speed and angular velocity commands $u$ to the inverse kinematics node, which converts these to motor commands.}
    \label{fig:pipeline_jon_plante}
\end{figure}
\textbf{Inspiration:}
This technique was developed to emulate the way in which a human might center a vehicle in a lane using the left and right lane boundaries as guides. More specifically, the driver may steer the vehicle towards the point at which the boundary lines interesect at the horizon. Alg.~\ref{alg:jon_plante} provides an algorithmic description of this approach.

\textbf{Image preprocessing:}
The pipeline (Fig.~\ref{fig:pipeline_jon_plante}) begins by filtering input images to find the white and yellow lane markings by applying several steps detailed in Alg.~\ref{alg:jon_plante}.
Following several preprocessing steps, the lanes are detected and projected into the horizon. 
Their point of intersection is used as the target direction and used to calculate an angle of deviation that is later used to control the robot. 

\begin{algorithm}
\caption{Method by Jon Plante}\label{alg:jon_plante}
\begin{algorithmic}[1]
\Procedure{Image preprocessing}{}
\State  Receive image $y$ and convert to gray scale, and threshold (Otsu's binarization~\cite{otsu1979threshold})
\State  Erode image with $3\times 3$ kernel and dilute image with $3\times 3$ kernel
\State Find connected components
\State Determine whether a connected area is white or yellow using color averages
\State Image resizing to remove top part of image
\State Find midpoints of white and yellow areas
\State Fit a line to both yellow and white area midpoints. Find center line of image intersecting both lines.
\State Calculate angle of deviation from vertical.
\EndProcedure
\Procedure{Control}{}
\State Check which case of angle deviation applies
\State Apply the preset command output given the angle deviation case  
\EndProcedure
\Procedure{Motor commands}{}
\State Convert control command into motor commands
\EndProcedure
\end{algorithmic}
\end{algorithm}

\textbf{Control:}
The desired steering angle and forward speed are read from a lookup table based on the current heading deviations.
Generally, a smaller angle corresponded to greater speed, since vehicles need to slow down in curves. 
When no yellow line is detected on the left side of the image, the robot assumes that it is not in the right lane and thus switches to another controller that pushes the Duckiebot to realign itself by going gradually to the right.

\textbf{Competition experience and lessons:}
The detailed preprocessing likely helped the Duckiebot drive around even in conditions that it was not originally tested on. However, during the competition, the approach might not have been flexible enough given the case-based controller. Likewise, lane markings were detected using averaging thresholds for white and yellow lines that are likely sensitive to the lighting conditions.

\subsubsection{Contribution from Participant 3: Vincent Mai}\label{sec:results-cont-3}

\begin{figure}[t]
    \centering
    \includegraphics[width=0.8\textwidth]{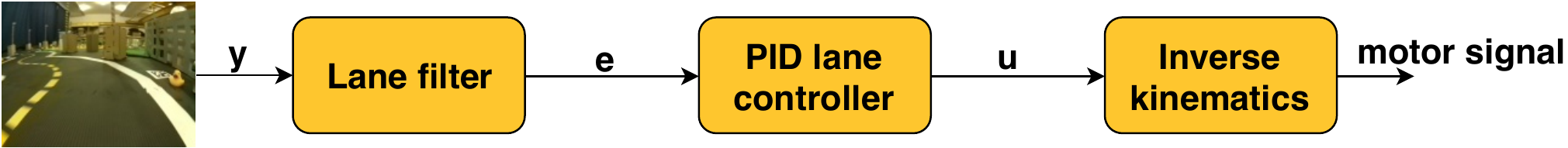}
    \caption{The ROS-based PID pipeline adapted by Vincent Mai. Images $y$ are processed by the lane detector to estimate position error $e$ for the PID controller. 
    The controller determines the speed and angular velocity commands $u$ and the inverse kinematics node converts these into motor commands.}
    \label{fig:pipeline_vincent_mai}
\end{figure}

% the idea + figure
\textbf{Inspiration:}
The rationale of this approach is to tune a tried and tested solution without reinventing the wheel. 
The participant started with the ROS-based lane following baseline~\cite{paull2017duckietown}. This method detects lane markings in the first-person images and uses these detections to estimate the Duckiebot's distance to the centerline of the lane. This error is fed to a PID controller. The participant identified several weaknesses of this pipeline, and modified several parameters to improve performance.

%the method, what was improved
\begin{algorithm}
\caption{Method by Vincent Mai}\label{alg:vincent_mai}
\begin{algorithmic}[1]
\Procedure{Lane filter}{}
\State Receive image $y$
\State Compensate illumination changes using clustering methods
\State Detect color of lanes and fit lines to lanes
\State Estimate position error $e$ to middle of right lane
\EndProcedure
\Procedure{PID controller}{}
\State Compute control signal $u$ using new and stored position error estimates
\EndProcedure
\Procedure{Motor commands}{}
\State Convert control command into motor commands
\EndProcedure
\end{algorithmic}
\end{algorithm}

\textbf{Lane filter:}
The image processing pipeline is prone to erroneous detections, labeling yellow lines as white lines (near the end of tiles) and green grass as yellow lane markings. The participant addressed this by manually tuning the HSV color space parameters. The method also reduced the original extent of the image crop to provide observations of lane markings that were useful for estimating deviation from the center line.

\textbf{Lane controller:}
The original setpoint of the PID controller positioned the Duckiebot too far to the right of the center line. The participant tuned this parameter to ensure a steady-state position closer to the middle of the right lane. In order to increase the Duckiebot's speed, the participant used greater gains on the commands to the motors. Further, the participant tuned the controller's derivative gain to reduce oscillations in the Duckiebot's heading.

% experiences: What was good, what was bad
\textbf{Competition experience and lessons:}
Starting with an existing baseline was helpful but also limited the design freedom of the method. Having access to a physical Duckiebot supported the tuning and development of this pipeline. However, the participants found that several tuned parameters were sensitive to the specific Duckiebot on which the pipeline was run as well as the environment. For example, the lighting conditions during the competition, which were different from those of the testing environment, introduced errors in lane detection. Additionally, the revised motor gains that were used to increase speed did not directly transfer to the Duckiebot used in the competition. The participant also found that control parameters tuned to optimize performance in simulation resulted in poor performance on the real robot.

\subsubsection{Contribution from Participant 4: Team JetBrains ---
Mikita Sazanovich, 
Oleg Svidchenko,
Aleksandra Malysheva,
Kirill Krinkin,
Maksim Kuzmin,
Aleksandr Semenov,
Vadim Volodin, and
Aleksei Shpilman}\label{sec:results-cont-4}

\begin{figure}[!t]
    \centering
    \includegraphics[width=0.8\textwidth]{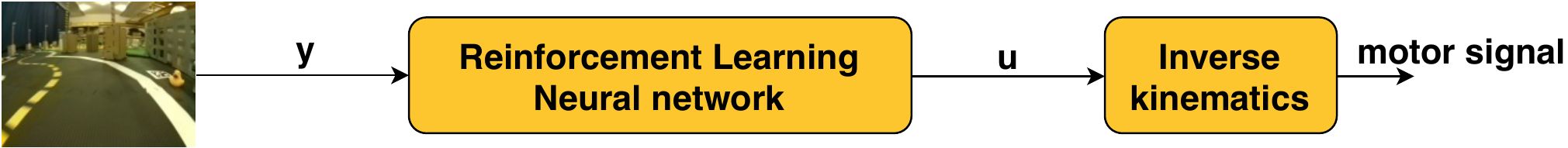}
    \caption{Reinforcement learning pipeline developed by team JetBrains. Images $y$ are processed by a neural network trained through reinforcement learning. 
    The network computes the speed and angular velocity commands $u$ and the inverse kinematics node converts these into motor commands.}
    \label{fig:jetbrains}
\end{figure}

\textbf{Inspiration:} The team was interested in applying deep reinforcement learning to solve for a lane-following policy. Further, they wanted to use this as an opportunity to explore the feasibility of transferring policies trained in simulation to the real world.

\begin{figure}[ht]
    \centering
    \includegraphics[width=0.6\textwidth]{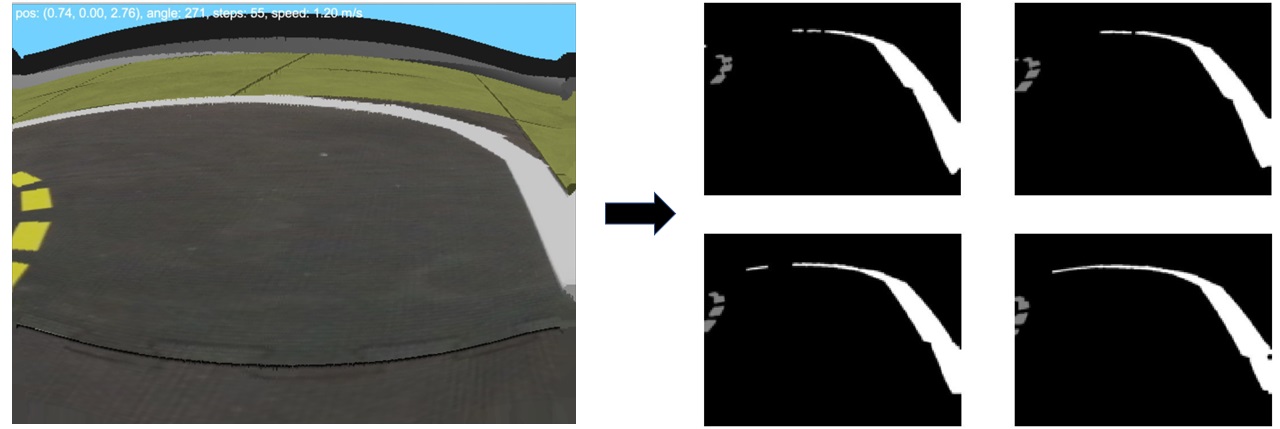}
    \caption{As a preprocessing step, filters for white and yellow colors were applied to the last four frames.}
    \label{fig:jetbrains-input}
\end{figure}

\textbf{Learning approach:}
The team used their framework for parallel deep reinforcement learning\footnote{Accessible online at \url{https://github.com/iasawseen/MultiServerRL}}. 
The neural network consisted of five convolutional layers (the first layer contained 32 $9\times 9$ filters, and the remaining four contained 32 $5 \times 5$ filters), followed by two fully connected layers (with 768 and 48 neurons, respectively). Input to the network consisted of the the last four frames downsampled to 120 by 160 pixels and filtered for white and yellow color as shown in Fig.~\ref{fig:jetbrains-input}. The network was trained using the Deep Deterministic Policy Gradient algorithm~\cite{lillicrap2015continuous}. 
The training was conducted in three stages: First, on a full track, then on the most problematic regions (identified by the largest cross-track error), and again on a full track. 
The reward had the following form: $[\text{current speed}] x [\cos(\text{target angle}- \text{current angle})]$. 
Angles were extracted in the simulator during training, but are not necessary for inference since the method only relies on preprocessed camera frames. 

\textbf{Competition experience and lessons:}
As most finalist teams, Team JetBrain found sim-to-real transfer to be difficult. It is not clear whether or not reinforcement learning techniques are more susceptible to this problem, however there are heuristic control algorithms that can be calibrated for real-world Duckietown and perform quite well.
The team plans to address this problem in future work, perhaps with a hybrid approach that uses reinforcement learning and machine learning not to devise the exact action, but to either preprocess environmental data or fine-tune parameters for standard control algorithms.

\subsubsection{Contribution from Participant 5: Team SAIC Moscow---Vladimir Aliev, Anton Mashikhin, Stepan Troeshestov and Sergey Nikolenko}\label{sec:results-cont-5}

\begin{figure}[ht]
    \centering
    \includegraphics[width=0.8\textwidth]{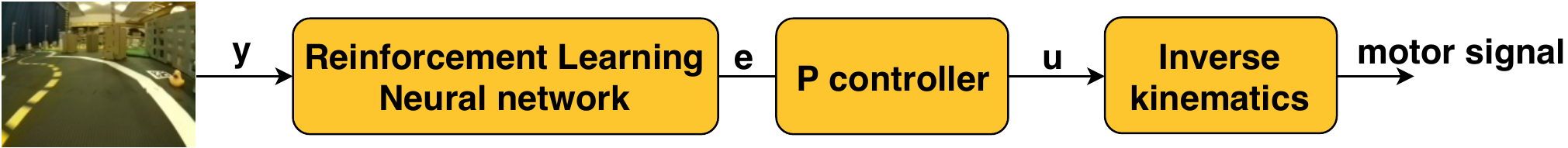}
    \caption{The reinforcement learning pipeline developed by Team SAIC Moscow. Images $y$ are processed by a neural network trained through reinforcement learning to output a score of how closely the heading aligns with the lane direction. 
    This signal is used as an error signal to a proportional controller which in turn computes $u$. The inverse kinematics node converts these commands into motor commands.}
    \label{fig:saic}
\end{figure}

\textbf{Inspiration:}
From the start of the challenge, the team decided to stick with an end-to-end learnable approach. 
The participants started with a simple Deep Deterministic Policy Gradient (DDPG)~\cite{lillicrap2015continuous} baseline that predicted steering angle and velocity from the input images. 
However, they found that this approach gave rather poor results with the Duckiebot only moving forward and backward. 
The team experimented with exploration strategies, reinforcement learning algorithms, and training parameters. They converged on an original exploration scheme and the distributed Twin Delayed DDPG~\cite{fujimoto2018addressing} learning algorithm.

\textbf{Image preprocessing:}
Input images were resized to $60 \times 80$ and converted to grayscale. Three consecutive frames were stacked along the channel dimension to provide temporal information.

\textbf{Network Architecture:}
The method used a convolutional neural network with five convolutional layers followed by two fully connected layers. 
The feature sizes for the convolutional layers gradually increased from 8 to 64, where the first layers had stride 2.
Fully connected layers had a hidden size of 300. 
The network inputs three-channel images and outputs the dot product between Duckiebot heading and center of the lane.

\textbf{Exploration and learning:}
A key contribution was to implement guided exploration, as it was hard for the agent to come up with a meaningful strategy during random exploration for this task.
The participants decided to add guidance using the default proportional controller provided by AI-DO.
To train either proportional controller or trained network were randomly selected to act. 
This technique greatly reduced the time required to train the agent and significantly improved the final policy quality. 
The team implemented distributed off-policy training where agents interact with the simulator and collect experience independently and afterwards send it to shared storage. 
The training workers executed the Twin Delayed DDPG algorithm~\cite{fujimoto2018addressing}. 
After a number of iterations, parameters of the network were broadcasted to the workers.

\textbf{Controller:}
The participants used a modified proportional controller to compute the motor commands. The controller took as input the dot product between the lane and robot heading as predicted by the network, and used a proportional term to drive the dot product close to one while moving at maximum speed.
Although hardcoding the maximum speed may seem controversial, it worked well in simulation.

\textbf{Competition experience and lessons:}
The team found the baseline algorithms that were provided to help in speeding up the development process.
However, they felt that access to physical hardware is critical to developing a navigation algorithm, particularly with learning-based algorithms that require additional work to transfer from simulation. 
The team implemented simple augmentations and used domain randomization in the simulator, but it proved to be insufficient to transfer the trained policy to real-world inputs. 
The domain transfer requires additional future work to allow deploying end-to-end learning-based models on real devices.

\subsection{Finals}\label{sec:results-finals}

\begin{figure}[ht]
    \centering
    \includegraphics[width=0.3\textwidth, height=3cm]{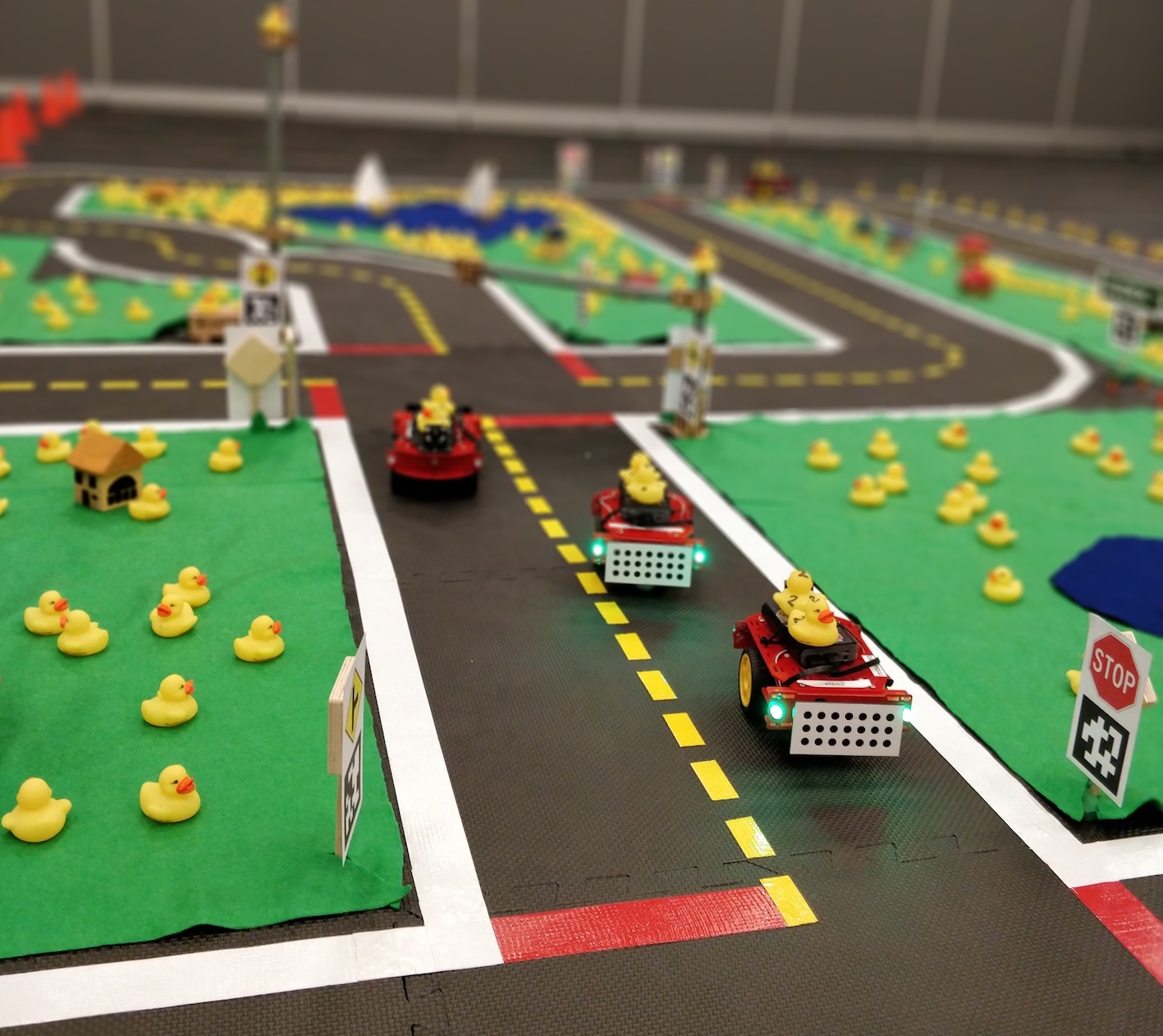}
    \includegraphics[width=0.3\textwidth, height=3cm]{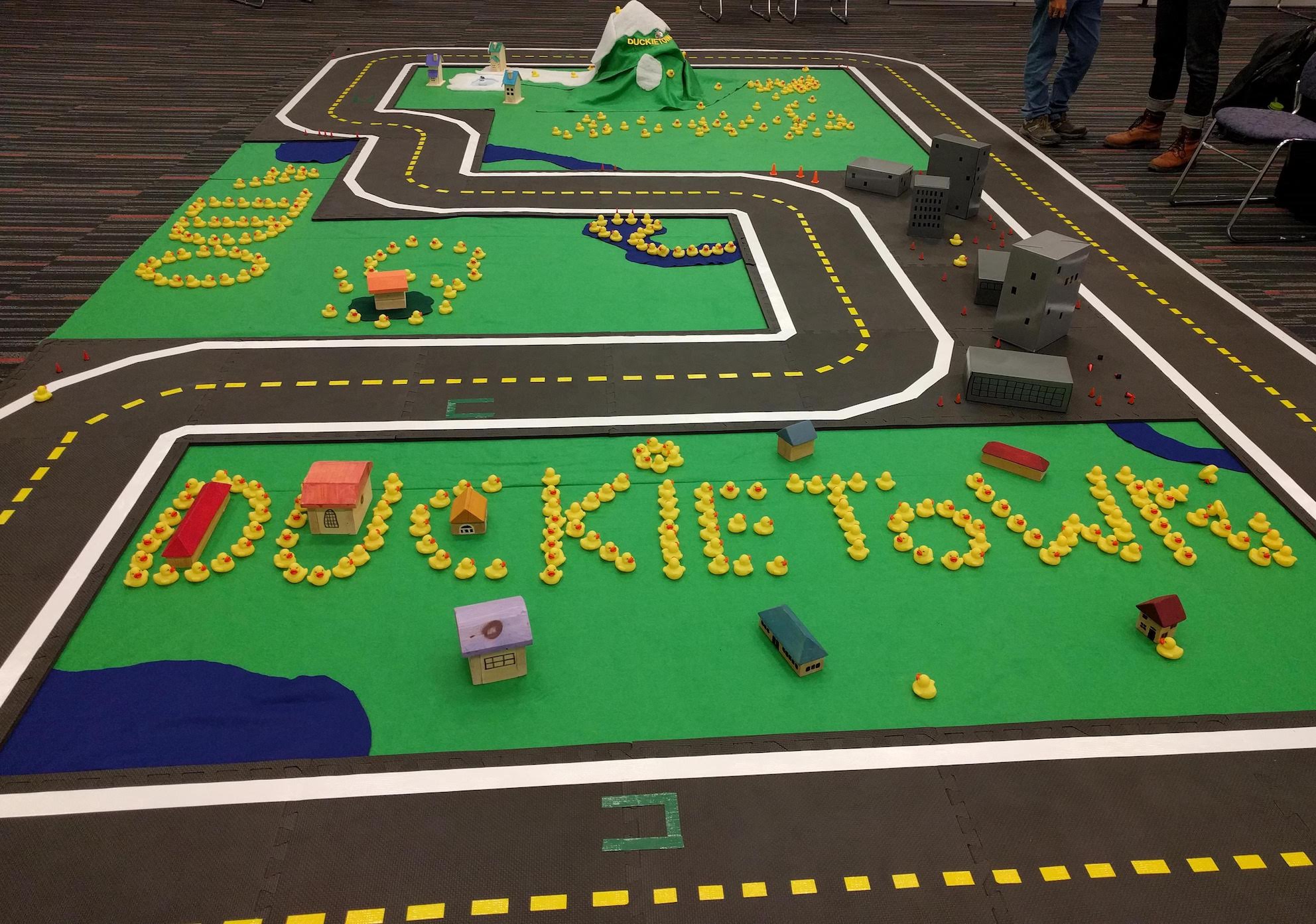}
    \includegraphics[width=0.3\textwidth, height=3cm]{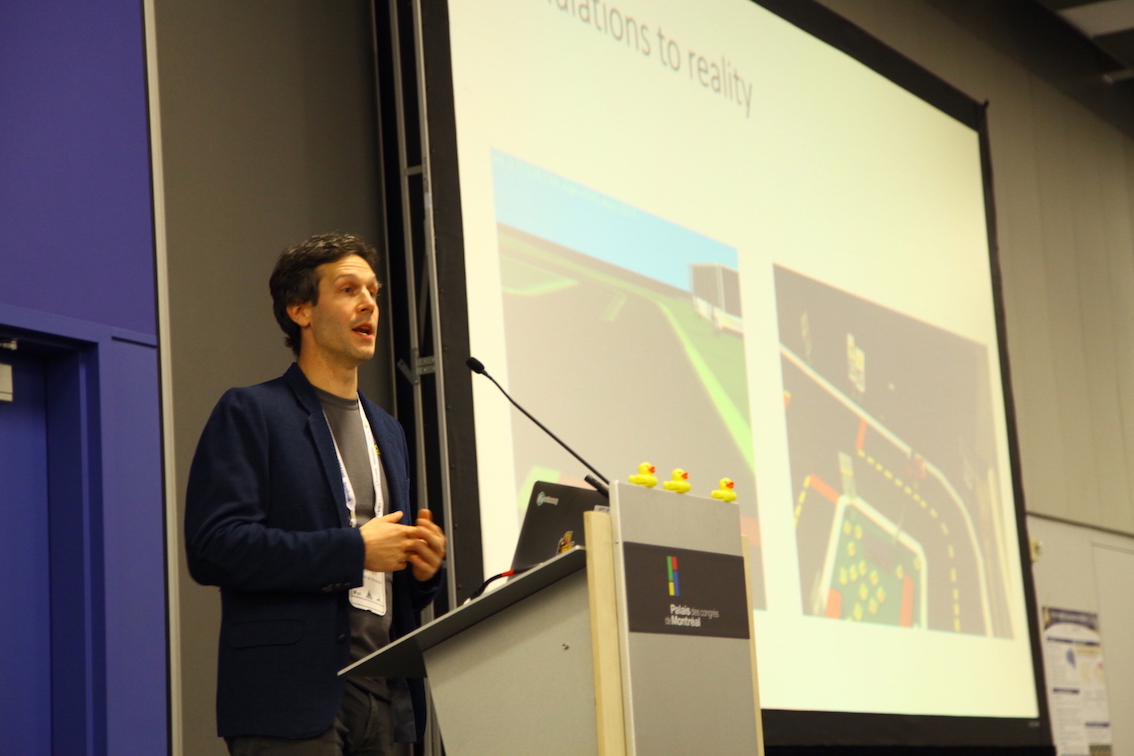}
    \caption{Impressions from the finals of AI-DO 1 at NeurIPS 2018}
\end{figure}
As the peak of the competition, the finals let the best five contestants in simulation compete on real Duckiebots at NeurIPS 2018. The
Duckiebots were placed in five starting positions spread out across the competition Duckietown. 
A judge assessed the performance of each robot in terms of the duration of time that it remained in the driveable area and the number of tiles that it traversed (Tab.~\ref{tab:results}).
The best submission of each finalist was uploaded onto our competition computer, which streamed images from the Duckiebot via WiFi. 
The algorithm then computed the desired actions and sent them back to the Duckiebot.

\begin{table*}[ht]
\centering
	 \caption{\emph{Results of the live competition in terms of the number of seconds before the vehicle exited the lane and the number of tiles that were traversed. Table entries denote the Duckiebot performance from five different starting positions as rounds. Participants were ranked in terms of distance traveled, with travel time breaking any ties.}}
	 \vspace{2mm}
	 {\small
	 \begin{tabular}{lcccccccc}
	   \toprule
	   \textbf{Contestant} &
	   \textbf{Round 1} &
  	   \textbf{Round 2} &  
 	   \textbf{Round 3} &
	   \textbf{Round 4}  &
	   \textbf{Round 5}  &
	   \textbf{Cumulative} 
       \\
	   \midrule
	   \emph{Wei Gao}        & 3.3 / 1 & 3.9 / 1 & 2.0 / 1 &  23.0 / 12 &  5.0 / 3  & \textbf{37 / \underline{18}}\\ % checks out
	   \emph{SAIC Moscow}   & 6.0 / 3  & 2.0 / 1 & 2.0 / 1 &   3.0 / 1  &  2.0 / 1  & 15 / \underline{7}\\ % checks out
	   \emph{Team JetBrains} & 16.0 / 2  & 1.0 / 0 & 4.0 / 1 &   0.0 / 0  &  8.0 / 1  & 29 / \underline{4}\\ % checks out
	   \emph{Jon Plante}    & 18.0 / 2 & 1.0 / 0 & 7.0 / 3 &   3.0 / 1  &  5.0 / 3  & 34 / \underline{9}\\ % checks out
	   \emph{Vincent Mai}   & 2.0 / 1  & 1.0 / 0 & 3.0 / 2 &  14.0 / 1  &  3.0 / 2 &  23/ \underline{9}\\ % checks out
	   \bottomrule
	 \end{tabular}}
	 \label{tab:results}
\end{table*}

\textbf{Outcomes:} 
Thanks to an exceptionally long run in round four, contestant Wei Gao won the AI-DO 1 live finals. 
Most runs however did not travel more than three tiles---likely due to differences between simulation and real Duckietown, such as lighting and the cheering crowd in the background during the competition. 
Additionally, as may be observed in Fig.~\ref{fig:driving_behavior}, some of the finalist submissions appear to have overfit to the simulation environment. 
We conclude that methods developed in simulation may have considerable difficulty in reaching the performance of methods developed on physical hardware. 

\textbf{Awards:} The finalists were awarded \$3000 worth of Amazon web services (AWS) credits and a trip to the nearest nuTonomy (Aptiv) location for a ride in one of their self-driving cars.

% ------------------------------------------
%% ANALYSIS SECTION:
\section{AI-DO 1: Lessons Learned}
\label{sec:lessons-learned}
Duckiebots were not the only entities that went through a learning process during AI-DO 1. 
Below we identify which aspects were beneficial and should be continued as part of subsequent competitions, and which should be revised. We separate the discussion into two categories: technical and non-technical.

% ----------------------------------------------------

\subsection{Technical}
The following reflects upon what we learned making and running AI-DO 1 from a technical point of view. 

\subsubsection{Successes} 

\textit{Software infrastructure}

The use of Docker allowed us to run submission code on many different computational platforms and operating systems. 
A central and beneficial feature of Docker is that any submission is reproducible, it can still be downloaded, tried and tested today. 
Likewise, it was helpful that all submissions were evaluated on a server yet could also be run locally. This allowed participants to rapidly prototype solutions locally in an environment that was an exact functional match of the evaluation environment. As a result, competitors could do many local evaluations and only submit an entry when they were confident that they would surpass previous results, which alleviated stress on our server infrastructure.

\textit{Simulation}

The custom simulator was used for initial benchmarking. Since it was developed in lightweight Python/OpenGL, it enabled rapid training (up to 1k frames/sec) for quick  prototyping of ML-based solutions.    
Online evaluations and comprehensive visualizations\footnote{Any submission is visualized on \url{https://challenges.duckietown.org/v3/} by clicking its submission number.}  were especially helpful in assessing performance and better understanding behavior.
The fact that the Duckietown environment is so structured and well-specified, yet simple, allowed for easy generation of conforming map configurations in simulation with relatively little effort. 

\textit{Baselines and templates} 

The containerized baselines and templates provided entry points that made it easy for competitors from different communities to begin working with the platform, essentially with just a few lines of code, as described in Sec.~\ref{subsec:baselines}. These baselines made the competition accessible to roboticists and the ML communities alike, by providing interfaces to the standard tools of each.  
These learning resources were reinforced further through Facebook live events and a question forum that explained the AI-DO software in more detail.

\subsubsection{To Be Improved}

\textit{Software infrastructure}

A large amount of effort was dedicated to the use of containerization via Docker. While containerization provides several benefits including reproducibility, the Docker infrastructure still poses challenges. For example, 
Docker changed versions in the middle of the competition wreaking temporary havoc to our dependencies. 

An awareness of, and strategy for, dealing with resource constraints are essential to robotic systems. However, these constraints still have to be implemented using Docker.
This gave an advantage to approaches that were computationally more expensive, when the inverse was desired. For example, a submission that ran on a Raspberry Pi should have been preferred over a submission that ran off-board on an expensive GPU.

\textit{Submission evaluation}

All submissions were evaluated on one local server, which made the submission infrastructure fragile. There were instances when participants were unable to make submissions because the server was down or we had too many submissions entered at the same time. In the future, we will aim to move evaluations to the cloud to offset these issues.

Enhanced debugging and evaluation tools would have made it considerably easier to diagnose issues during the local evaluation phase. Users could be shown a richer set of diagnostics and logs from the simulator directly after their submission was evaluated, or from recorded playback logs of an on-Duckiebot trial. 

All submissions were publicly visible. Competitors were able to download, view, and re-submit other images using the public hash code. While we are not aware of any participants accessing another team's submissions during the competition, we intend to change the submission infrastructure to make the process more secure.

Furthermore, evaluations were not statistically significant since the simulations considered 20 episodes in only 1 map. Subsequent competitions will include more maps and more variability in the simulated environments as well as more evaluation episodes. 

Bringing submissions tested in simulation to testing on real robots is a major component of AI-DO. While we provided ample opportunity for testing in simulation in AI-DO 1, there was insufficient opportunity for competitors to test on the hardware, which was apparent in the quality of the runs in the live event. 

\textit{Simulation}

While the finals of AI-DO 1 proved to be a good demonstration that the simulator-reality gap is large, we did not have a metric for quantifying it. 
The Duckietown codebase has had a working lane following solution since 2016 \cite{paull2017duckietown}, so we know that a robust solution to the LF challenge was feasible. 
However, in simulation this robust solution did not perform well. In contrast, solutions that did perform well in simulation did not work on the real hardware. Since most competitors did the majority of their development in the simulation environment, this resulted in sub-optimal performance at the live event. 

We \textit{must} provide better tools for competitors to be able to bridge the reality gap more easily. These tools should include: better API to modify and/or randomize parameters in the simulator, easier access to real robot logs, and easier access to real robot infrastructure either remotely (through robotariums) or locally.

\clearpage

\textit{Baselines and templates}

Duplication in the boilerplate code resulted in diversity in the interface among various templates and baselines. We plan to reduce this duplication to make it easier to change protocols between different challenges. 

\textit{Logs}

Logs represent valuable data from which to train algorithms, yet we are unaware of any competitor that made use of the $\sim$16 hours of logs that were made available. Our conclusion was that the log database was not sufficiently advertised and that the workflow for using the logs in a learning context was not explicit.

\subsection{Non-Technical}

The following discusses the non-technical aspects of AI-DO 1. 

\subsubsection{Successes} 

\textit{Competition logistics}

The Duckietown platform~\cite{paull2017duckietown} has proven to be an ideal setup for benchmarking algorithms as it is simple, highly structured and formally specified, yet modular and easily customised. It is also accessible to competitors as the robot and city hardware is readily available and inexpensive. 

\textit{Available resources}

The Duckietown platform contains a rich assortment of learning resources in the Duckiebook~\cite{Duckiebook} and on the Duckietown website~\cite{duckietown_website}, as well as a large body of open-source software for running Duckiebots within Duckietown. The wealth of existing resources reduced the technical support demand on the organizers.

\textit{Engagement}

\begin{figure}[tb]
    \centering
    \includegraphics[width=0.4\textwidth, height=3.5cm]{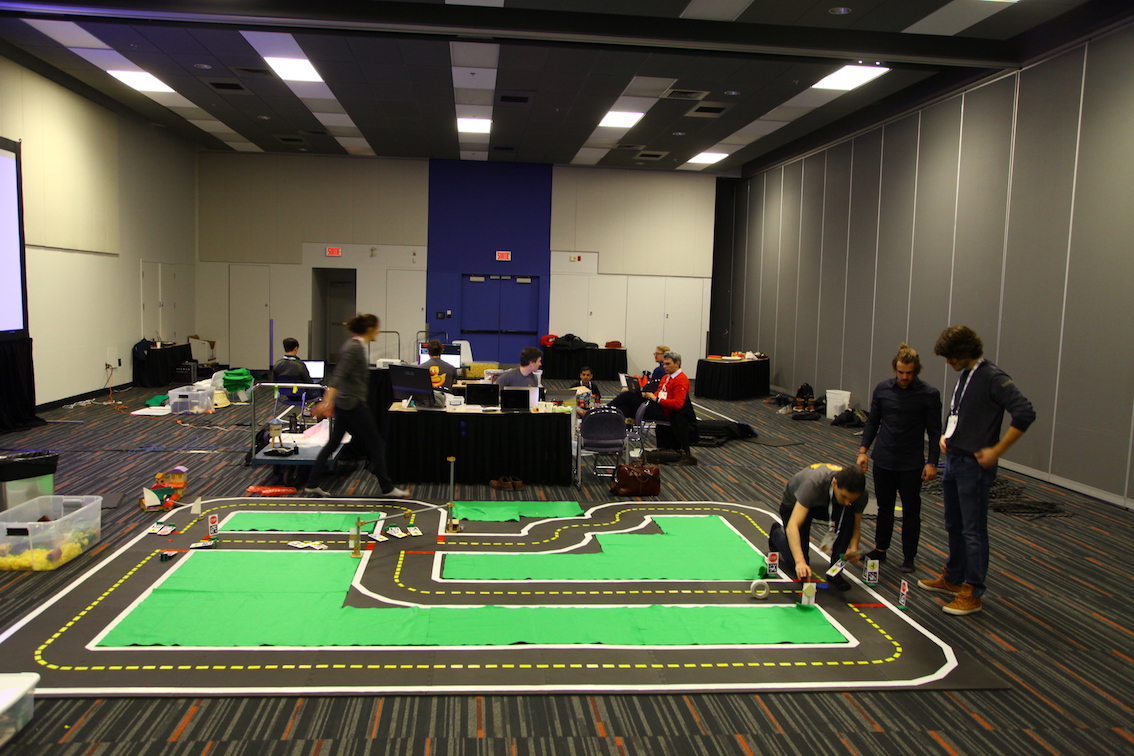}
    \includegraphics[width=0.4\textwidth, height=3.5cm]{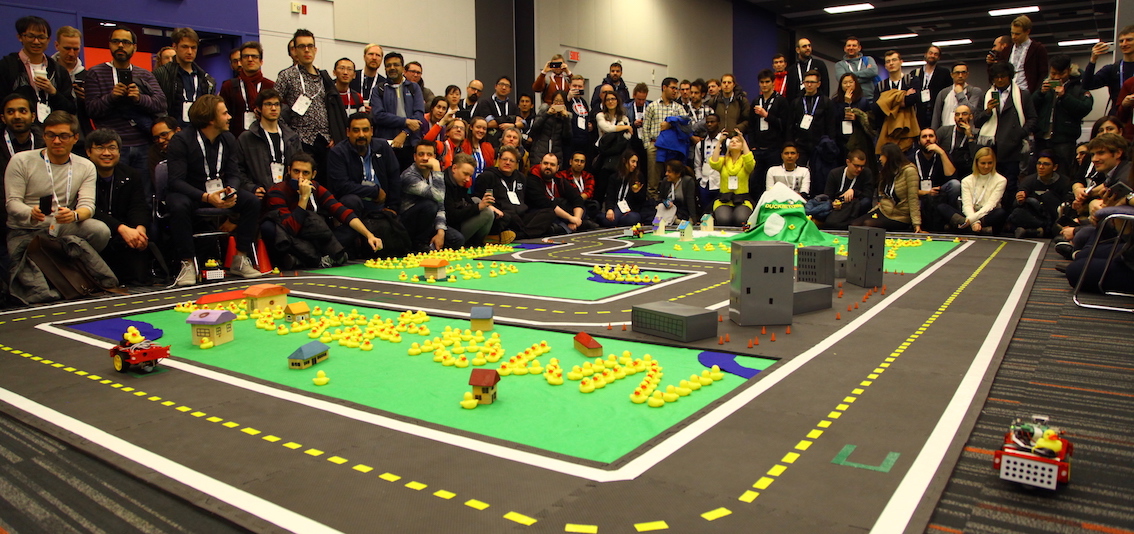}
    \caption{The AI-DO 1 event. \textbf{Left:} The setup required an entire day and a small army of volunteers. \textbf{Right:} Audience at the finals of AI-DO 1 at NeurIPS 2018 in Montr\'eal, Canada. For applause, the public squeezed rubber duckies instead of clapping.
    }
    \label{fig:public_involvement}
\end{figure}

% Online resources: Many baselines were available, duckiebook, demos, facebook live events to introduce to a topic
Especially positive was the public's emotional reception at the AI-DO 1 finals, as shown in Fig.~\ref{fig:public_involvement}-right. 
Both the Duckie-theme and competition were well-received. Instead of clapping between rounds the audience squeezed yellow rubber duckies.

\subsubsection{To Be Improved}

Similar to many larger projects, AI-DO 1 fell victim to the \emph{planning fallacy}~\cite{planning_fallacy} wherein we were more optimistic in our timeline than our general experience should have suggested. 

\textit{Competition logistics}

The amount of setup required for the 1.5 hour event was significant. We required access to the competition space for the entire day  and needed to move a significant amount of materials into the competition venue (see Fig.~\ref{fig:public_involvement}-left). Making this process smoother and simpler would be an important step towards making the competition easily repeatable.

From a broader perspective there was little community involvement in the development of the AI-DO infrastructure. 

\textit{Rules}

\begin{figure}[tb]
    \centering
    \includegraphics[width=\textwidth]{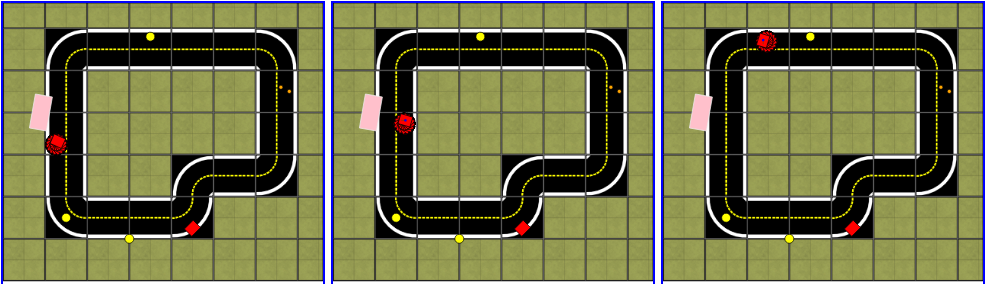}
    \caption{Submission to Challenge LFV by user ``sodobeta''. Example of degenerate solution found by reinforcement learning as a result of poorly specified evaluation metrics. Submission originally got a high score for ``Traveled Distance'' even though it is just spinning in a circle. 
    }
    \label{fig:spinning}
\end{figure}

As the competition infrastructure and our understanding of the tasks evolved, the rules evaluating competitors changed too frequently and too close to the end of the competition. This may have made it harder to participate easily and improve consistently. 
It was surprisingly difficult to define the evaluation metrics in a way that promoted the kind of behavior we sought. Reinforcement learning algorithms are well-known for finding these types of loopholes. As one example, the evaluation of the ``Traveled Distance'' did not consider progress down the lane but only linear speed at first. As a result, it was possible to get a high score by spinning in circles as fast as possible (Fig.~\ref{fig:spinning}). 

% ------------------------------------------
%% CONCLUSION/OUTLOOK SECTION:
\section{Next steps: AI-DO 2 at ICRA}
\label{sec:icra}

Armed with the experience of the first version of the competition, we have set out to improve some of the key difficulties and to introduce some innovations.

\textit{Rules and evaluation}

Early on in the preparation for AI-DO 2, we will fix the rules and how participants will be evaluated. 
Special care will be placed on using a simple evaluation procedure that nevertheless captures the essence of the approach. 

A big change that we are in dire need of is the introduction of hidden tests that are a standard best practice in ML. 
There will thus be a test set that is not visible to any participant.
Additionally, even in the visible evaluations, there will be more variability in submissions, such as different environments, lighting conditions or other types of randomization. 

To facilitate the development of better driving approaches, we will make it easier to use logs of past driving behavior and to take logs and make them available. 

Finally, we will investigate running submission on demand in the cloud which should increase our flexibility and, while we did not have any serious issues, make the AI-DO 2 submission procedure reliable. 

\textit{Software infrastructure}

Another step in improving the competition is to make use of the functionality of Docker to place resource constraints on submissions such as available computational speed and memory. 

An innovative new aspect we will introduce are \emph{modules}. 
Modules are functions that compute part of a larger pipeline and may be used in a plug-and-play fashion. 
We envision that these modules are themselves a ``micro-competition'' in the sense that better modules may lead to better submissions to the AI-DO tasks. 
Again, these modules would be benchmarked with respect to computational and memory requirements. 

\textit{Simulation} 

The simulator will continue to be an integral part of the competition. 
Making it more realistic by providing better and more maps, more realistic physics and visualization and domain randomization will be decisive steps towards AI-DO 2. 

As part of including resource constraints into the workflow, we plan to include two types of evaluation.  
\begin{itemize}
\item Fixed-length time steps between actions independent of computational load
\item Variable-length time steps as long as it takes to make a decision
\end{itemize}

\textit{Evaluation on actual hardware}

The largest improvement for experimental evaluation will be the introduction of Robotariums. 
These are Duckietowns with automatic localization of Duckiebots and subsequent automatic scoring of their behavior. 
Having access to Robotariums is the key to giving participants valuable feedback on how their solutions work on real hardware.

\textit{Outreach and engagement}

We hope to attract many more competitors. 
The guiding philosophy should be to limit the necessary work of participants to follow their own ideas while giving them the opportunity to venture beyond by contributing back to the Duckietown and AI-DO ecosystem. 

\section{Conclusion}
\label{sec:conclusion}

We have presented an overview of the AI Driving Olympics. The competition seeks to establish a benchmark for evaluating machine learning and ``classical'' algorithms on real physical robots. The competition leverages the Duckietown platform: a low-cost and accessible environment for autonomous driving research and education. 
We have described the software infrastructure, tools, and baseline algorithm implementations that make it easy for robotics engineers and machine learning experts alike to compete. 

The first instance of the AI-DO, AI-DO 1, was held in conjunction with  NeurIPS in Dec. 2018. We have outlined some of the successes and weaknesses of this first experience. We will leverage this information to improve future competitions, starting with the AI-DO 2, which will be held at ICRA in 2019.

For science to advance we need a reproducible testbed to understand which approaches work and what their trade-offs are. 
We hope that the AI Driving Olympics can serve as a step in this direction in the domain of self-driving robotics. 
Above all we want to highlight the unique design trade-offs of embodied intelligence due to resource constraints, non-scalar cost functions and reproducible robotic algorithm testing.

\section*{Acknowledgements} We would like to thank NeurIPS and in particular Sergio Escalera and Ralf Herbrich for giving us the opportunity to share the AI Driving Olympics with the machine learning community. We are grateful to Amazon AWS and Aptiv for their sponsorship and hands-on help that went into this competition. 
We are also grateful to  the many students in Montr\'eal, Zurich, Taiwan, Boston, Chicago, and many others that have shaped Duckietown and AI-DO into what they are today. 
% ------------------------------------------

\bibliographystyle{plainnat}
\bibliography{bib.bib}

\end{document}